\documentclass{article}

\usepackage{arxiv}

\usepackage[utf8]{inputenc} 
\usepackage[T1]{fontenc}    
\usepackage{hyperref}       
\usepackage{url}            
\usepackage{booktabs}       
\usepackage{amsfonts}       
\usepackage{nicefrac}       
\usepackage{microtype}      
\usepackage{lipsum}

\usepackage{graphicx}
\usepackage{multirow}%
\usepackage{amsmath,amssymb,amsfonts}%
\usepackage{amsthm}%
\usepackage{mathrsfs}%
\usepackage[title]{appendix}%
\usepackage{xcolor}%
\usepackage{textcomp}%
\usepackage{manyfoot}%
\usepackage{algorithm}%
\usepackage{algorithmicx}%
\usepackage{algpseudocode}%
\usepackage{listings}%
\usepackage{subcaption}
\usepackage{siunitx}  

\theoremstyle{thmstyleone}%

\theoremstyle{thmstyletwo}%
\theoremstyle{thmstylethree}%

\usepackage[backend=biber, style=numeric, sorting=none]{biblatex}
\graphicspath{ {./images/} }

\title{Smooth-Distill: A Self-distillation Framework for Multitask Learning with Wearable Sensor Data}

\author{
 Hoang-Dieu Vu \\
  Faculty of EEE, Phenikaa School of Engineering\\
  Phenikaa University\\
  Yen Nghia, Hanoi 12116, Vietnam \\
  Graduate University of Science and Technology, VAST \\
  Hanoi 122300, Vietnam \\
  \texttt{dieu.vuhoang@phenikaa-uni.edu.vn} \\
   \And
Duc-Nghia Tran \\
  Institute of Information Technology\\
  VAST\\
  Hanoi 122300, Vietnam \\
  \texttt{nghiatd@ioit.ac.vn}
   \And   
 Quang-Tu Pham \\
  Faculty of EEE, Phenikaa School of Engineering\\
  Phenikaa University\\
  Yen Nghia, Hanoi 12116, Vietnam \\
  \texttt{20010752@st.phenikaa-uni.edu.vn} \\
  \And
 Hieu H. Pham \\
  College of Engineering \& Computer Science\\
  and VinUni-Illinois Smart Health Center\\
  VinUniversity\\
  Hanoi 10000, Vietnam \\
  \texttt{hieu.ph@vinuni.edu.vn} \\
\And
     Nicolas Vuillerme \\
  AGEIS, Université Grenoble Alpes\\
  Grenoble 38000, France \\
  Institut Universitaire de France\\
  Paris 75005, France \\
  \texttt{nicolas.vuillerme@univ-grenoble-alpes.fr} \\
\And    
Duc-Tan Tran \\
  Faculty of EEE, Phenikaa School of Engineering\\
  Phenikaa University\\
  Yen Nghia, Hanoi 12116, Vietnam \\
  \texttt{tan.tranduc@phenikaa-uni.edu.vn} \\
}

\begin{document}
\maketitle
\begin{abstract}
This paper introduces Smooth-Distill, a novel self-distillation framework designed to simultaneously perform human activity recognition (HAR) and sensor placement detection using wearable sensor data. The proposed approach utilizes a unified CNN-based architecture, MTL-net, which processes accelerometer data and branches into two outputs for each respective task. Unlike conventional distillation methods that require separate teacher and student models, the proposed framework utilizes a smoothed, historical version of the model itself as the teacher, significantly reducing training computational overhead while maintaining performance benefits. To support this research, we developed a comprehensive accelerometer-based dataset capturing 12 distinct sleep postures across three different wearing positions, complementing two existing public datasets (MHealth and WISDM). Experimental results show that Smooth-Distill consistently outperforms alternative approaches across different evaluation scenarios, achieving notable improvements in both human activity recognition and device placement detection tasks. This method demonstrates enhanced stability in convergence patterns during training and exhibits reduced overfitting compared to traditional multitask learning baselines. This framework contributes to the practical implementation of knowledge distillation in human activity recognition systems, offering an effective solution for multitask learning with accelerometer data that balances accuracy and training efficiency. More broadly, it reduces the computational cost of model training, which is critical for scenarios requiring frequent model updates or training on resource-constrained platforms. The code and model are available at https://github.com/Kuan2vn/smooth\_distill. 
\end{abstract}


\section{Introduction}
The rapid advancement of wearable technologies has revolutionized healthcare monitoring and human activity recognition (HAR) applications \cite{Wearable_device2, Wearable_device3}. The integration of machine learning techniques with sensor data processing has enabled sophisticated monitoring systems capable of tracking physical activity, sedentary behavior, and sleep, as well as other health metrics \cite{Wearable_device_focus_deep_side, Wearable_device_deep}. However, the effectiveness of these systems extends beyond the mere availability of sensor data, as contextual factors significantly influence data quality and interpretation accuracy \cite{Context_sensor_data}.

Location for sensor placement on the body emerges as a particularly critical contextual factor that directly affects data quality and system performance \cite{position_HAR_reliable, Fixed_position_HAR}. Different sensor placement positions tend to yield varying performance outcomes across distinct application domains. Ankle-mounted sensors are commonly preferred for step counting and gait analysis applications \cite{ankle_application, ankle_comparision2}, as they typically demonstrate favorable results in capturing lower-limb movement patterns \cite{ankle_comparision, ankle_comparision2}. Hip and lower back positions are frequently selected for general activity recognition tasks \cite{hips_comparasion}, with researchers often reporting satisfactory performance in detecting diverse daily activities from these locations \cite{hips_application}. Wrist-worn devices are generally favored for sleep quality monitoring applications, where they tend to provide reliable data for sleep pattern analysis \cite{wrist_application, wrist_application2, wrist_application3}. This position-dependent variation in sensor performance characteristics is of utmost importance for real-world deployment of HAR systems. Traditional approaches that rely on fixed sensor placements may encounter practical limitations, as the effectiveness of specific placements varies considerably across different application contexts and for various targeted end-users, potentially constraining the broader applicability of such systems.

To address these limitations, researchers have pursued two primary strategies: developing deep learning models that maintain robust performance across different sensor placement location on the body (position-independent approaches) \cite{DNN_position_independent} and constructing multitask learning frameworks that simultaneously perform HAR and sensor position recognition \cite{Multitask_HAR_position_FedOpenHAR, Multitask_HAR_position2}. The latter approach, which forms the foundation of this research, leverages the inherent relationship between sensor placement and activity patterns to enhance overall system performance.

Multitask Learning (MTL) represents a machine learning paradigm wherein a single shared model is trained to simultaneously learn multiple related tasks \cite{first_multitask, base_multitask}. The fundamental objective of MTL is to exploit useful information and shared underlying patterns between related tasks to improve the generalization performance of all involved tasks. This is achieved by leveraging both commonalities and differences between tasks through the optimization of multiple objectives simultaneously, typically by minimizing a composite loss function represented as a weighted sum of individual task-specific loss functions. The advantages of MTL include enhanced data generalization, greater efficiency in data processing, reduced overfitting through shared representations across tasks, and decreased computational demands compared to deploying multiple independent models for each task.

Despite these advantages, MTL presents inherent challenges, particularly the risk of negative transfer and task interference \cite{multitask_negative, multitask_negative2}. The task selection problem in MTL represents a critical bottleneck, as weakly related or conflicting tasks can lead to negative transfer effects \cite{negative_transfer}. For HAR and sensor position recognition, this implies that a strong fundamental relationship between the two tasks must exist for MTL to be effective, or sophisticated strategies must be employed to mitigate complexity. Existing research confirms a manageable relationship between these tasks, suggesting that multitask learning is a viable strategy.

\begin{figure*}[t]%
\centering
\includegraphics[width=0.8\textwidth]{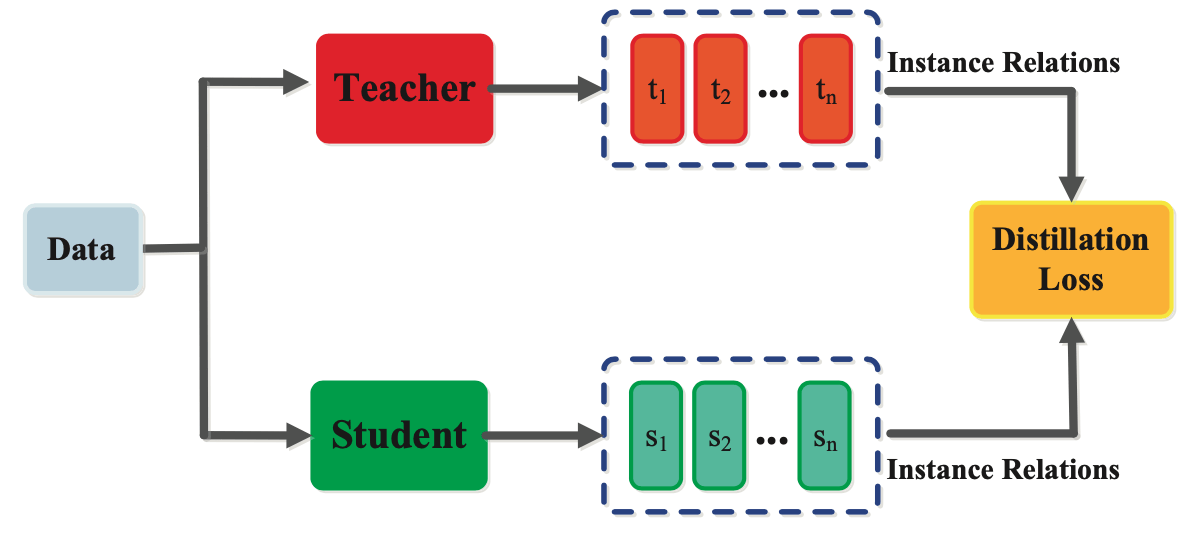}
\caption{Standard knowledge distillation setup, where a student model learns from both ground-truth labels and the soft predictions of a pre-trained teacher model \cite{KD_Figure}.}
\label{fig:knowledge_distillation_framework}
\end{figure*}

Self-distillation techniques offer promising solutions to address limitations of traditional multitask learning approaches \cite{Self_distill_multitask, Self_distill_multitask_SDMTCNN}. Knowledge distillation, originally introduced as a method for transferring knowledge from a large teacher model to a smaller student model by leveraging soft probability distributions as supervision signals \cite{Knowledgedistillation}, has evolved to encompass self-distillation frameworks where a model serves as its own teacher. An overview of the standard knowledge distillation process is illustrated in Figure~\ref{fig:knowledge_distillation_framework}. 

In the context of multitask learning, self-distillation eliminates the need for separate teacher-student architectures while providing significant regularization benefits. The technique mitigates negative transfer effects by enabling the model to learn from its own stable predictions, thereby reducing task interference and improving convergence stability. This approach demonstrates particular effectiveness in scenarios where tasks exhibit varying levels of difficulty or learning rates, as the self-distillation mechanism helps balance learning dynamics across different objectives while maintaining consistent performance throughout the training process \cite{Self_distill_multitask, Self_distill_multitask_SDMTCNN}.

However, conventional distillation methods require pre-trained teacher models, substantially increasing training time and computational costs \cite{Pham2022Distill, hajimoradlou2022selfsupervised, felix2023selfdistilledrepresentationlearningtime}. To address this limitation, we propose a novel self-distillation method called "Smooth-Distill," wherein the teacher model represents a historical, smoothed, and stabilized version of the student model itself. This approach substantially reduces computational requirements by eliminating the need for separate teacher training while maintaining comparable performance improvements, making it particularly suitable for resource-constrained wearable computing environments.

To support this research, we have constructed a comprehensive accelerometer dataset capturing 12 distinct sleep postures across three different device wearing positions, complementing two existing datasets. This newly developed dataset serves as the foundation for our multitask deep learning model, which incorporates our proposed novel self-distillation technique. The model is designed to simultaneously predict sleep postures or activities and to identify device placement locations, demonstrating the practical application of our Smooth-Distill approach in a real-world context.

The main contributions of this work are summarized as follows:

\begin{itemize}
    \item We propose Smooth-Distill, a novel and computationally efficient self-distillation framework, in which the model is trained to match a smoothed version of its own historical parameters. This approach eliminates the need for a separate teacher model and is particularly suitable for multitask learning scenarios.
    
    \item We design and implement MTL-net, a convolutional neural network tailored for human activity recognition (HAR). The model is capable of jointly predicting both activity classes—including a comprehensive set of sleep postures—and sensor placements, using only raw accelerometer data.
    
    \item We construct and publicly release a new dataset for sleep posture recognition, comprising annotated data collected from 23 participants, covering 12 distinct postures across three sensor placements. To the best of our knowledge, this is one of the first datasets to offer such granularity in sleep-related activity recognition.
    
    \item We conduct extensive experimental evaluations on three datasets: the newly proposed Sleep dataset, MHealth, and WISDM. Results demonstrate that our approach consistently outperforms strong baselines, including single-task models, conventional multitask frameworks, and existing self-distillation strategies, in terms of both accuracy and generalization.
\end{itemize}

The remainder of this paper is organized as follows: Section \ref{preliminaries} (Preliminaries) introduces the problem formulation in the context of multitask learning, outlines conventional self-distillation approaches, and presents the key idea behind our proposed Smooth-Distill method. Section \ref{sec:method} (Methodology) then provides a detailed description of each component within the Smooth-Distill framework. Section \ref{sec:experiment} (Experiment) describes the datasets used, the preprocessing procedures, and the comparative methods for evaluation. Section \ref{sec:results} presents the experimental results and key findings. Section \ref{sec:discussion} discusses the benefits, limitations, and potential applications of our proposed approach. Finally, Section \ref{sec:conclusion} concludes the paper by summarizing main contributions, addressing open challenges, and outlining directions for future research.

\section{Preliminaries}
\label{preliminaries}

We leverage self-distillation algorithms to enhance model performance, which offers several key benefits including improved accuracy, better generalization capabilities, and enhanced model robustness. Self-distillation techniques have demonstrated their versatility across diverse domains, including multitask learning scenarios with sensor data processing. In the following sections, we first establish the foundational concepts of multitask learning and examine the conventional self-distillation approach in Section \ref{preli:conventional_sd}. This traditional methodology has proven its effectiveness in improving model performance while facilitating faster convergence during training. However, the requirement for a well-pretrained teacher model makes this approach computationally expensive, as it necessitates training the model twice, potentially doubling the overall training time. To address these limitations, we propose a novel approach to self-distillation specifically designed for multitask training in Section \ref{preli:smooth_distill}, where the teacher model functions as a smoothed, historical representation of the student model. This innovation eliminates the need to train a separate teacher model independently from the student, significantly reducing computational costs and training time compared to conventional self-distillation methods.

\subsection{Multitask Learning and Conventional Self-Distillation Framework}
\label{preli:conventional_sd}

\subsubsection*{Multitask Learning Framework}

The problem formulation centers on a dataset denoted as $\mathcal{D} = \{(x_i, y_i^{(1)}, y_i^{(2)})\}_{i=1}^{N}$ containing $N$ samples, where each input sample $x_i$ corresponds to two distinct output labels $y_i^{(1)}$ and $y_i^{(2)}$. The input data comprises windowed time-series information from sensor measurements, specifically acceleration data with consistent sensor orientations utilized in this research. The first output task involves predicting sleep postures (such as prone, supine, lateral positions, etc.) or general activity recognition (such as running, walking, standing, lying, etc.). The second output task focuses on predicting sensor placement locations, encompassing ankle, chest, wrist and abdomen positions.

The learning objective involves training a deep learning model parameterized by weights $\theta$ to minimize an appropriately defined loss function. The optimization problem is mathematically formulated as:
\[
\theta^* = \arg\min_\theta \mathcal{L}(\theta)
\]

In multitask learning scenarios, a prevalent approach involves combining the individual loss functions from each task. This research adopts the strategy of integrating cross-entropy losses from both tasks through a weighting parameter $\alpha$ that balances the relative importance of the two tasks:
\begin{equation}
\mathcal{L}(\theta) = \alpha \cdot \mathcal{L}_{\text{CE}}^{(1)} + (1 - \alpha) \cdot \mathcal{L}_{\text{CE}}^{(2)} \label{eq:multitask_loss}
\end{equation}
This formulation ensures that both tasks contribute appropriately to the overall learning objective while maintaining flexibility in task prioritization through the weighting mechanism.

\subsubsection*{Conventional Self-Distillation Framework}
\label{sec:conventional_born_again_sd}

Self-distillation represents a knowledge transfer paradigm where a well-trained "teacher" model imparts its learned knowledge to a "student" model, typically resulting in improved performance for the student network. This approach leverages the insight that teacher models can provide richer supervisory signals beyond simple ground-truth labels.

The baseline implementation examined in this research follows the Born-Again Self-Distillation methodology \cite{born_again_SD} as outlined in Algorithm \ref{alg:born_again_sd}. This conventional approach operates through a two-stage training process. Initially, a model undergoes standard training using the multitask loss function described in equation \eqref{eq:multitask_loss} until convergence. Subsequently, this trained model assumes the role of teacher for a second training phase, where knowledge distillation techniques are employed.

\begin{algorithm}[t]
\caption{Born-Again Self-Distillation (Single-Task Setting)}
\label{alg:born_again_sd}
\begin{algorithmic}[1]
\Require Initial parameters $\theta_0$; learning rate $\eta$; temperature $\tau$; distillation weight $\lambda$
\State \textbf{Stage 1: Train teacher model}
\State Initialize model: $\theta_T \gets \theta_0$
\For{epoch = 1 to $E_T$}
    \For{each minibatch $(x, y) \in \mathcal{D}$}
        \State $z_T \gets \text{Teacher}(x; \theta_T)$ \Comment{Logits}
        \State Compute cross-entropy loss: $\mathcal{L}_{\text{CE}}$
        \State Update teacher model: $\theta_T$
    \EndFor
\EndFor

\Statex

\State \textbf{Stage 2: Train student model with distillation}
\State Initialize student model: $\theta_S \gets \theta_0$
\For{epoch = 1 to $E_S$}
    \For{each minibatch $(x, y) \in \mathcal{D}$}
        \State $z_S \gets \text{Student}(x; \theta_S)$ \Comment{Student logits}
        \State $z_T \gets \text{Teacher}(x; \theta_T)$ \Comment{Teacher logits}
        \State Compute cross-entropy loss: $\mathcal{L}_{\text{CE}}$
        \State Compute distillation loss using Eq.~\eqref{eq:distillation_loss}: $\mathcal{L}_{\text{Distill}}$
        \State Compute total loss using Eq.~\eqref{eq:total_BA_loss}: $\mathcal{L}_{\text{total}}$
        \State Update student model: $\theta_S$
    \EndFor
\EndFor
\end{algorithmic}
\end{algorithm}

The knowledge transfer mechanism typically operates through optimization of the distributional differences between teacher and student model predictions. The standard distillation loss utilizes Kullback-Leibler divergence to measure and minimize these differences:
\begin{equation}
\mathcal{L}_{\text{Distill}} = \tau^2 \cdot \frac{1}{N} \sum_{n=1}^{N} \sum_{k=1}^{C} p_{n,k}^T(\tau) \cdot \left( \log p_{n,k}^T(\tau) - \log p_{n,k}^S(\tau) \right) \label{eq:distillation_loss}
\end{equation}
Where:
\begin{itemize}
    \item $p_{n,k}^T(\tau) = \text{softmax}(z_{T,n,k} / \tau)$ is the softened probability distribution over classes from the teacher model.
    \item $\log p_{n,k}^S(\tau) = \log\text{-softmax}(z_{S,n,k} / \tau)$ is the log of softened probabilities from the student model.
    \item $N$ is the total number of training samples.
    \item $C$ is the number of classes.
    \item $\tau$ is the temperature parameter controlling the softness of the distributions.
\end{itemize}

The complete training objective in self-distillation combines both the supervised multitask loss and the distillation loss. A typical formulation incorporates a distillation weight hyperparameter $\lambda$ to balance the influence of knowledge distillation versus direct supervision from ground-truth labels. Specifically, the total loss is defined as:

\begin{equation}
\mathcal{L}_{\text{total}} = (1 - \lambda) \cdot \mathcal{L}_{\text{CE}} + \lambda \cdot \mathcal{L}_{\text{Distill}} \label{eq:total_BA_loss}
\end{equation}

where $\mathcal{L}_{\text{CE}}$ denotes the cross-entropy loss computed using ground-truth labels, and $\mathcal{L}_{\text{Distill}}$ refers to the Kullback-Leibler divergence between the softened output distributions of the teacher and student models. The hyperparameter $\lambda \in [0, 1]$ balances the contributions from direct supervision and knowledge distillation.

\subsubsection*{Challenges and Proposed Direction}

The conventional self-distillation framework demonstrates substantial effectiveness in achieving performance improvements across various datasets and applications. The approach benefits from the enhanced supervisory signal provided by teacher models, which often contain richer information than simple one-hot ground-truth labels. Additionally, the framework provides a systematic methodology for knowledge transfer that has proven robust across different architectures and domains.

However, this conventional approach presents significant computational limitations. The requirement to train two separate models sequentially---first the teacher, then the student---substantially increases both training duration and computational resource demands. This inefficiency becomes particularly pronounced in resource-constrained environments or when dealing with large-scale datasets and complex model architectures.

To address these computational challenges while preserving the benefits of knowledge distillation, this research proposes an innovative approach that eliminates the need for separate teacher-student training phases, thereby reducing computational overhead while maintaining or potentially improving upon the performance gains achieved through conventional self-distillation methods.

\subsection{A Novel Self-Distillation Approach for Multitask Training}
\label{preli:smooth_distill}

To address the substantial increase in training time and computational cost associated with training separate teacher models, we propose the Smooth-Distillation method. This approach aims to achieve significantly improved recognition performance for both human activity and sensor placement classification tasks, without substantially increasing computational costs during training.

Instead of requiring a pre-trained teacher model, our technique creates an adaptive teacher that evolves alongside the student model. The core innovation lies in maintaining a smoothed parameter version of the student model itself, which serves as a stable reference point for knowledge transfer. This dynamic teacher provides increasingly refined guidance throughout the training process while eliminating the need for separate teacher training phases.

Specifically, at each time step t, the teacher model parameters are updated through a weighted average of their previous values and the current student model parameters:

\begin{equation}
    \theta^{\text{T}}_t = \beta \cdot \theta^{\text{T}}_{t-1} + (1 - \beta) \cdot \theta^{\text{S}}_t
    \label{smooth_teacher}
\end{equation}

where $\theta^{\text{T}}_t$ represents the teacher parameters at time step $t$, $\theta^{\text{S}}_t$ represents the student model parameters at time step $t$, and $\beta$ is the smoothing coefficient controlling the averaging rate. We employ $\beta = 0.999$ to ensure the teacher model captures stable representations while gradually incorporating new information.

This approach completely eliminates the separate teacher training phase, transforming the traditional two-stage training process into a single, unified training procedure that achieves the benefits of knowledge distillation without the associated computational overhead.

\subsubsection*{Notations and Definitions} 
To clarify the formulas and the proposed method, we present the key notations used throughout the paper. These include parameters of the Student and Teacher models, loss functions for multitask training and knowledge distillation, and hyperparameters governing the training process. Table \ref{tab:key_notations} below provides a comprehensive summary of these notations along with their descriptions.

\begin{table}[h]
\centering
\caption{Summary of Key Notations.}
\renewcommand{\arraystretch}{1.2} 
\label{tab:key_notations}
\begin{tabular}{c m{7cm}} 
\toprule
\textbf{Notation} & \multicolumn{1}{c}{\textbf{Description}} \\ 
\midrule
$(x, y^{(i)})$ & Input sample ($x$) and its ground-truth label for task $i$ ($y^{(i)}$). \\
$\theta^{\text{S}}_t, \theta^{\text{T}}_t$ & Student and Teacher model parameters at time step $t$. \\
$\mathcal{L}_{\text{CE}}^{(i)}$ & Cross-Entropy loss for task $i$. \\
$\mathcal{L}_{\text{Distill}}$ & Knowledge distillation loss. \\
$\mathcal{L}_{\text{total}}$ & Total training loss for the student model. \\
$\alpha$ & Weighting parameter to balance multitask losses. \\
$\tau$ & Temperature parameter for softening distributions in distillation. \\
$\lambda$ & Weighting parameter to balance supervised loss and distillation loss. \\
$\beta$ & Smoothing coefficient for updating the teacher parameters. \\
\bottomrule
\end{tabular}
\end{table}

\section{Methodology}
\label{sec:method}

\subsection{Framework Overview}
\label{method:overall_framework}

The proposed system presents a comprehensive multitask learning approach that simultaneously predicts sleep postures and sensor placement locations through a unified neural network architecture. The framework leverages the MTL-net model, a CNN-based architecture specifically designed for time series analysis, to process acceleration sensor data and generate dual classification outputs.

\begin{figure*}[t]%
\centering
\includegraphics[width=1.0\textwidth]{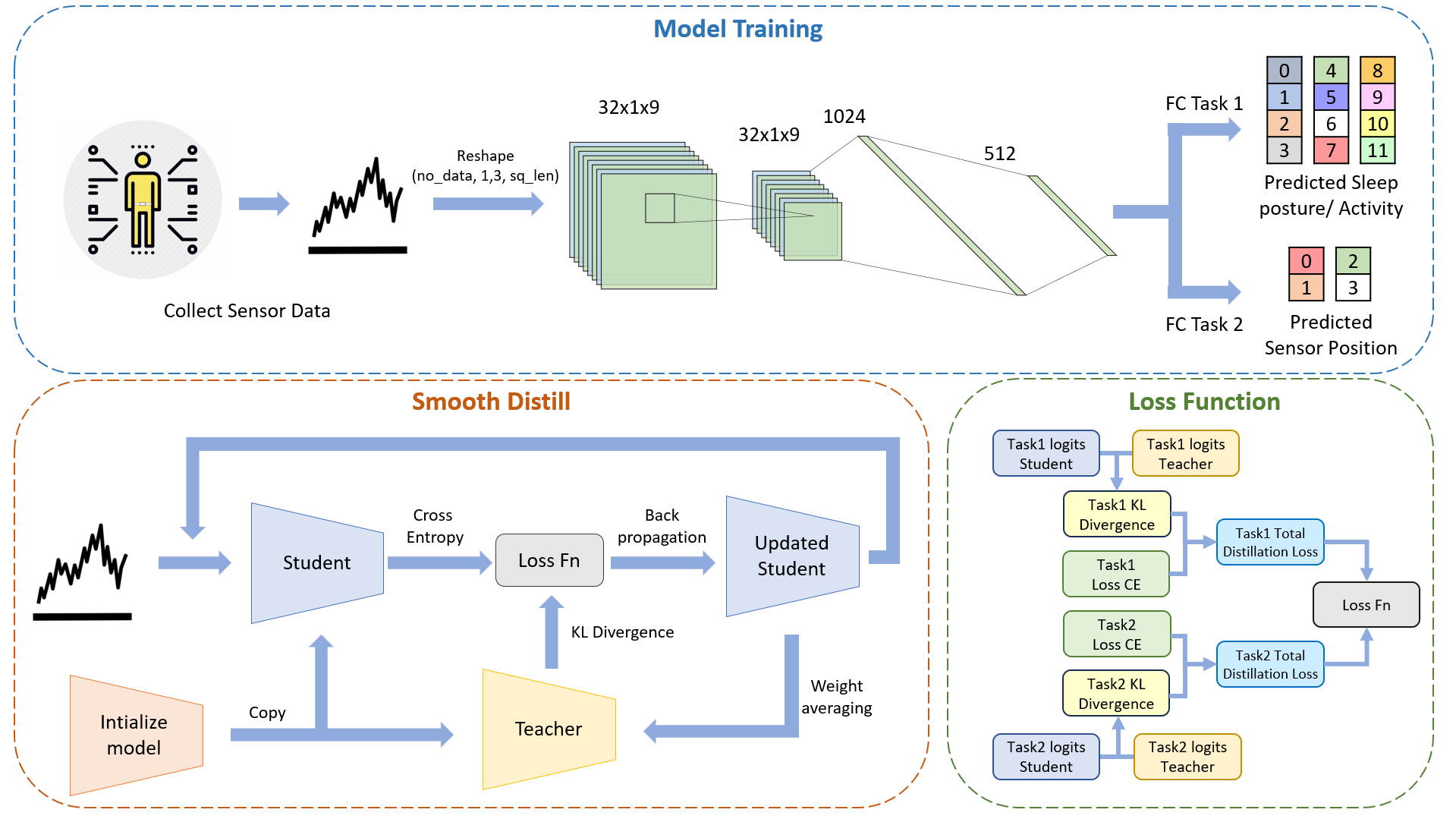}
\caption{Overview of the proposed smooth-distill framework, consists of three main components: (Top) Standard model training pipeline: collected sensor data is fed into a deep learning model to predict sleep posture/activity and sensor placement. (Bottom left) Core Smooth-Distill mechanism: student and teacher model outputs are used to compute a combined distillation loss, which is backpropagated to update the student. The updated student parameters are then used to update the teacher via weighted parameter averaging. (Bottom right) Detailed breakdown of the combined distillation loss components, comprising task-specific cross-entropy loss for classification accuracy and Kullback-Leibler (KL) divergence between student and teacher logits to ensure effective knowledge transfer from teacher to student model.}
\label{fig:abstract}
\end{figure*}

Figure \ref{fig:abstract} illustrates the complete processing pipeline, demonstrating how input sensor data flows through the MTL-net architecture with integrated student-teacher interactions. The system employs a smooth distillation mechanism that facilitates knowledge transfer between related tasks, enabling the model to learn shared representations while maintaining task-specific outputs. The loss computation process incorporates both individual task losses and cross-task knowledge distillation losses to optimize the overall model performance.

The framework encompasses three key components that will be detailed in subsequent sections. The MTL-net architecture (Section \ref{method:model_architecture}) provides the foundational neural network structure for feature extraction and multitask learning. The Smooth-Distill (Section \ref{method:smooth_distill}) mechanism facilitates effective knowledge transfer between tasks through controlled distillation processes. Finally, the multitask loss function (Section \ref{method:loss_formulation}) combines individual classification objectives with regularization terms to ensure balanced learning across all prediction targets.

This integrated approach enables the system to achieve enhanced performance on both sleep posture recognition and sensor placement detection while maintaining computational efficiency through shared feature representations.

\subsection{Model Architecture}
\label{method:model_architecture}

In this research, we propose a convolutional neural network architecture specifically designed for time-series data analysis, which we term MTL-net (Multitask Learning-Net). This architecture was developed to address the specific requirements of processing temporal data from wearable sensors with enhanced versatility and performance.

MTL-net is a convolutional neural network specifically engineered for efficient feature extraction from time-series data. The architecture leverages convolutional operations to detect spatial and temporal patterns within sensor data streams. The network employs a series of convolutional layers that progressively extract hierarchical features from raw accelerometer data, complemented by pooling operations that reduce dimensionality while preserving essential information. The architecture concludes with fully connected layers that interpret the extracted features for final classification tasks. Batch normalization layers are integrated throughout the network to stabilize the learning process and improve convergence. The comprehensive architecture design, including specific layer configurations and parameters, is detailed in Figure \ref{fig:abstract}.

The distinguishing feature of MTL-net lies in its multitask learning capability, which enables simultaneous prediction of multiple related tasks. Specifically, our implementation facilitates concurrent prediction of both sensor position and posture recognition, thereby enhancing the model's practical utility and versatility in real-world applications. This multitask approach allows the network to leverage shared representations across related tasks, potentially improving overall performance while maintaining computational efficiency.

The MTL-net architecture demonstrates robust performance in extracting meaningful features from accelerometer data across various sensor placements and activities. This capability makes it particularly well-suited for applications requiring comprehensive analysis of human movement patterns, including sleep posture recognition and activity classification tasks that form the core objectives of this research.

\subsection{The Smooth-Distill Mechanism}
\label{method:smooth_distill}

Building upon the self-distillation approach introduced in Section \ref{preli:smooth_distill}, the Smooth-Distill mechanism operationalizes the dynamic interaction between student and teacher models throughout the multitask training process. As established, both models are initialized with identical weights, creating the foundation for the adaptive teacher evolution that eliminates computational overhead associated with separate teacher training phases.

During training, the student model undergoes standard backpropagation updates based on the comprehensive loss function detailed in Section \ref{method:loss_formulation}. Following each student optimization step, the teacher model parameters are updated according to Equation \ref{smooth_teacher}, where the smoothing coefficient $\beta = 0.999$ ensures gradual incorporation of new information while maintaining stability. This weighted averaging mechanism transforms the traditional two-stage distillation process into a unified training procedure that achieves knowledge transfer benefits without additional computational cost.

The teacher model's inherent stability emerges from the weighted averaging update strategy, which effectively smooths out the fluctuations characteristic of standard gradient descent optimization. This stability enables the teacher to provide consistent soft targets that guide the student's learning beyond hard label matching, encouraging the development of robust feature representations that generalize effectively across different data distributions and multitask objectives.

\subsection{Multitask Loss Formulation}
\label{method:loss_formulation}

The comprehensive loss function that governs the entire learning process combines both cross-entropy losses and distillation losses across multiple tasks, formulated as:

\begin{equation}
\mathcal{L}_{\text{total}} = \alpha \cdot \left( \mathcal{L}_{\text{CE}}^{(1)} + \lambda \cdot \mathcal{L}_{\text{Distill}}^{(1)} \right) + (1 - \alpha) \cdot \left( \mathcal{L}_{\text{CE}}^{(2)} + \lambda \cdot \mathcal{L}_{\text{Distill}}^{(2)} \right)
\label{eq:total_loss}
\end{equation}

The cross-entropy components $\mathcal{L}_{\text{CE}}^{(1)}$ and $\mathcal{L}_{\text{CE}}^{(2)}$ represent the standard classification losses for each respective task, computed between the student model's output probabilities and the corresponding ground-truth labels. These components ensure that the model maintains accuracy on the primary classification objectives while participating in the knowledge distillation process.

The distillation losses $\mathcal{L}_{\text{Distill}}^{(1)}$ and $\mathcal{L}_{\text{Distill}}^{(2)}$,  formally defined by Equation \ref{eq:distillation_loss}, measure the Kullback-Leibler divergence between the student and teacher probability distributions, encouraging the student to replicate the teacher's confidence patterns across all classes rather than focusing solely on hard predictions. The temperature parameter $\tau$ within these distillation losses controls the softness of the probability distributions, with higher values producing more informative soft targets that emphasize relative relationships between different class predictions.

The task balancing coefficient $\alpha$ determines the relative importance between Task 1 and Task 2 within the overall optimization objective, while the distillation weight $\lambda$ controls the influence of knowledge transfer relative to standard supervised learning. These hyperparameters enable fine-tuned control over both task prioritization and the degree of internal knowledge distillation, allowing the framework to adapt to different problem requirements and data characteristics.

\subsection{The Complete Algorithm}
\label{method:algorithm}

Algorithm \ref{alg:smooth_distill} presents the comprehensive training procedure that integrates all components of the Smooth-Distill framework into a cohesive optimization process. The algorithm requires several key input parameters including the initial model parameters, learning rate, smoothing coefficient for teacher updates, temperature parameter for distillation, distillation weight, and task balancing coefficient.

\begin{algorithm}[t]
\caption{Smooth-Distillation for Multitask Learning}
\label{alg:smooth_distill}
\begin{algorithmic}[1]
\Require Initial parameters $\theta_0$; learning rate $\eta$; smoothing coefficient $\beta$; temperature $\tau$; distillation weight $\lambda$; task weight $\alpha$
\State Initialize student and teacher models: $\theta_S \gets \theta_0$, $\theta_T \gets \theta_0$
\For{epoch = 1 to $E$}
    \For{each minibatch $(x, y^{(1)}, y^{(2)}) \in \mathcal{D}$}
        \State $z_S^{(1)} \gets \text{Student}^{(1)}(x; \theta_S)$ \Comment{Task 1 student logits}
        \State $z_S^{(2)} \gets \text{Student}^{(2)}(x; \theta_S)$ \Comment{Task 2 student logits}
        \State $z_T^{(1)} \gets \text{Teacher}^{(1)}(x; \theta_T)$ \Comment{Task 1 teacher logits}
        \State $z_T^{(2)} \gets \text{Teacher}^{(2)}(x; \theta_T)$ \Comment{Task 2 teacher logits}
        \State Compute CE losses: $\mathcal{L}_{\text{CE}}^{(1)}$, $\mathcal{L}_{\text{CE}}^{(2)}$
        \State Compute distillation losses using Eq. \ref{eq:distillation_loss}: $\mathcal{L}_{\text{Distill}}^{(1)}$, $\mathcal{L}_{\text{Distill}}^{(2)}$
        \State Compute total loss using Eq. \ref{eq:total_loss}: $\mathcal{L}_{\text{total}}$
        \State Update student model: $\theta_S$
        \State Update teacher model using Eq. \ref{smooth_teacher}: $\theta_T$
    \EndFor
\EndFor
\end{algorithmic}
\end{algorithm}

The main training loop processes minibatches of input data, computing forward passes through both student and teacher models to generate task-specific logits for both classification objectives. The algorithm then calculates the individual cross-entropy losses and distillation losses according to Equation \ref{eq:distillation_loss} for each task before combining them through the total loss formulation presented in Equation \ref{eq:total_loss}. Following loss computation, the student model parameters are updated through Adam optimization, while the teacher model receives its weighted averaging update to maintain stability and accumulated knowledge.

This algorithmic framework demonstrates the seamless integration of multitask learning with self-distillation, creating an efficient training process that leverages shared representations while maintaining task-specific performance. The iterative nature of the algorithm ensures that both student and teacher models evolve together, with the teacher providing increasingly refined guidance as training progresses, ultimately leading to enhanced generalization capabilities across both classification tasks.

\section{Experiment}
\label{sec:experiment}
 
In this section, we present the experimental setup used in our study. We begin with a description of the datasets used and the evaluation metrics employed to assess model performance (Section \ref{experiment:dataset}). Next, we describe the preprocessing steps applied to the datasets, followed by an overview of the comparison methods, including our proposed approach and baseline techniques, as well as specific implementation details (Section \ref{experiment_design}).

\subsection{Datasets and Evaluation Metrics}
\label{experiment:dataset}
\subsubsection{Dataset Overview}
In this study, three datasets are utilized, including a sleep posture dataset developed by our team and two widely used human activity recognition (HAR) datasets featuring a broad range of physical activities. The diversity of postures and movements across these datasets provides a solid foundation for real-world activity recognition research.
\paragraph{Sleep dataset}

The Sleep Multitask dataset was specifically developed for this research to facilitate the analysis of body movement acceleration in twelve sleep postures. This dataset was designed to collect accelerometer data from 24 healthy young adult volunteers with diverse demographic profiles from Phenikaa University. Data collection utilized ADXL345 \cite{ADXL345} wearable sensors positioned on participants' chest, neck, and abdomen, capturing acceleration data at a consistent sampling frequency of 50 Hz across all sensor locations. The accelerometer was configured to operate at the ±2g range, which proved advantageous for sleep posture monitoring by enabling clear differentiation between sleep positions through enhanced resolution.

This study was approved by the Local Research Ethics Committee of Phenikaa University (LREC reference number: 024.22/DHP-HDDD). The 24 volunteers recruited for participation ranged in age from 19 to 29 years with a mean age of $22.08 \pm 1.82$ years. Their physical characteristics were systematically recorded, including height ($167.00 \pm 7.75$ cm), weight ($58.96 \pm 10.65$ kg), and BMI ($20.87 \pm 2.95$ kg/m$^2$). All participants received comprehensive instructions on the measurement procedures and sleep positioning requirements prior to data collection.

The dataset encompasses 12 distinct sleep postures measured on a standard bed, focusing on four fundamental positions: supine (0°), prone (270°), right lateral (90°), and left lateral (180°). Intermediate angles were also recorded, as detailed in Table \ref{tab:label_distribution}, which maps sleep positions based on the stomach's angle relative to the x-axis. Data acquisition was conducted in a structured manner, with each sleeping posture recorded for a duration of one minute. Individual data files were systematically labeled using a consistent naming convention that included both the participant identifier and the corresponding sleep position.

\begin{table}[t]
\small
\centering
\caption{Class distribution of sleep positions for n=24 participants in Sleep Dataset.}
\renewcommand{\arraystretch}{1.1} 
\label{tab:label_distribution}
\begin{tabular}{|c|c|c|c|c|c|}
\hline
\textbf{Label} & \textbf{Sleep postures} & \textbf{Abbreviation} & \textbf{Angle (degree)} & \textbf{Train set} & \textbf{Test set} \\ \hline
1 & Up (Supine)            & U    & 90   & 3574               & 894              \\
2 & Up Right               & UR   & 60   & 3563               & 891              \\
3 & Right Up               & RU   & 30   & 3542               & 885              \\
4 & Right (Lateral Right) & R    & 0    & 3617               & 905              \\
5 & Right Down             & RD   & 330  & 3600               & 900              \\
6 & Down Right             & DR   & 300  & 3532               & 883              \\
7 & Down (Prone)           & D    & 270  & 3621               & 906              \\
8 & Down Left              & DL   & 240  & 3498               & 874              \\
9 & Left Down              & LD   & 210  & 3578               & 894              \\
10 & Left (Lateral Left)    & L   & 180  & 3633               & 908              \\
11 & Left Up                & LU   & 150  & 3578               & 895              \\
12 & Up Left                & UL   & 120  & 3500               & 875              \\ \hline
\multicolumn{4}{|c|}{\textbf{Total}} & 42836              & 10710             \\
\multicolumn{4}{|c|}{\textbf{Mean}}  & 3569.67            & 892.50           \\
\multicolumn{4}{|c|}{\textbf{Median}} & 3576.00            & 894.00           \\ 
\multicolumn{4}{|c|}{\textbf{Min}}   & 3498               & 874              \\ 
\multicolumn{4}{|c|}{\textbf{Max}}   & 3633               & 908              \\ 
\multicolumn{4}{|c|}{\textbf{Standard deviation}}   & 44.91               & 11.45             \\ \hline
\end{tabular}
\end{table}

\paragraph{MHealth dataset}
The MHealth dataset, originating from the University of Granada (UGR), Spain \cite{Mhealth}. This dataset was designed to capture body motion and vital signs from 10 healthy volunteers with diverse profiles while performing 12 distinct physical activities.

Data collection employed Shimmer2 \cite{Shimmer2} wearable sensors positioned on the chest, right wrist, and left ankle. These sensors gathered multiple data types including acceleration, rate of turn (gyroscope readings), magnetic field orientation (magnetometer data), and two-lead ECG from the chest sensor; for the present study, only the accelerometer data was utilized. All sensors operated at a consistent sampling frequency of 50 Hz.

The 12 activities documented in the dataset comprise static postures (standing, sitting, lying), physical activities (walking, climbing stairs, cycling, jogging, running, jumping), and specific exercises (waist bends, frontal elevation of arms, and knees bending), with their detailed distribution shown in Table \ref{tab:Mhealth_data_distribution}. Data collection occurred in an out-of-lab environment without rigid constraints on activity execution. This approach enhances the dataset's generalizability to common daily activities through the diversity of body parts engaged and the varying intensity and speed of activity performance.

\begin{table}[t]
\small
\centering
\caption{Class distribution of sleep positions for n=10 participants in MHealth Dataset.}
\renewcommand{\arraystretch}{1.1} 
\label{tab:Mhealth_data_distribution}

\begin{tabular}{|c|c|c|c|}
\hline
\textbf{Label} & \textbf{Activity}           & \textbf{Train set} & \textbf{Test set} \\ \hline
1              & Null                        & 34899              & 8725              \\
2              & Standing still              & 1228               & 307               \\
3              & Sitting and relaxing        & 1229               & 307               \\
4              & Lying down                  & 1229               & 307               \\
5              & Walking                     & 1228               & 307               \\
6              & Climbing stairs             & 1230               & 308               \\
7              & Waist bends forward         & 1134               & 283               \\
8              & Frontal elevation of arms   & 1178               & 295               \\
9              & Knees bending (crouching)   & 1174               & 294               \\
10             & Cycling                     & 1229               & 307               \\
11             & Jogging                     & 1228               & 307               \\
12             & Running                     & 1229               & 308               \\
13             & Jump front \& back          & 413                & 103               \\ \hline
\multicolumn{2}{|c|}{\textbf{Total}} & 48628              & 12158             \\
\multicolumn{2}{|c|}{\textbf{Mean}}  & 3740.62            & 935.23            \\
\multicolumn{2}{|c|}{\textbf{Median}} & 1228.00            & 307.00            \\ 
\multicolumn{2}{|c|}{\textbf{Min}}   & 413                & 103               \\ 
\multicolumn{2}{|c|}{\textbf{Max}}   & 34899              & 8725              \\ 
\multicolumn{2}{|c|}{\textbf{Standard deviation}}   & 9364.57            & 2341.20           \\ \hline
\end{tabular}

\end{table}

\paragraph{WISDM dataset}

The WISDM (Wireless Sensor Data Mining) dataset \cite{wisdm_smartphone_and_smartwatch_activity_and_biometrics_dataset__507} was sourced from an activity recognition project involving 51 participants, primarily healthy young adults. Each individual performed 18 different physical activities for three minutes per activity while wearing a smartwatch (LG G Watch) and carrying a smartphone (Nexus 5, Nexus 5X, or Galaxy S6), with their detailed distribution also shown in Table \ref{tab:WISDM_distribution}. Both devices recorded accelerometer and gyroscope data at a sampling frequency of 20 Hz.

\begin{table}[t]
\small
\centering
\caption{Class distribution of sleep positions for n=51 participants in WISDM Dataset.}
\renewcommand{\arraystretch}{1.1} 
\label{tab:WISDM_distribution}

\begin{tabular}{|c|c|c|c|}
\hline
\textbf{Label} & \textbf{Activity}           & \textbf{Train set} & \textbf{Test set} \\ \hline
1              & Walking                     & 6537               & 1634              \\
2              & Jogging                     & 6322               & 1580              \\
3              & Stairs                      & 6170               & 1542              \\
4              & Sitting                     & 6368               & 1592              \\
5              & Standing                    & 6482               & 1621              \\
6              & Typing                      & 6019               & 1505              \\
7              & Brushing Teeth              & 6377               & 1595              \\
8              & Eating Soup                 & 6408               & 1602              \\
9              & Eating Chips                & 6286               & 1571              \\
10             & Eating Pasta                & 6038               & 1509              \\
11             & Drinking from Cup           & 6677               & 1669              \\
12             & Eating Sandwich             & 6262               & 1566              \\
13             & Kicking (Soccer Ball)       & 6510               & 1628              \\
14             & Playing Catch w/Tennis Ball & 6431               & 1608              \\
15             & Dribbling (Basketball)      & 6469               & 1617              \\
16             & Writing                     & 6350               & 1588              \\
17             & Clapping                    & 6358               & 1589              \\
18             & Folding Clothes             & 6354               & 1589              \\ \hline
\multicolumn{2}{|c|}{\textbf{Total}} & 114418             & 28605             \\
\multicolumn{2}{|c|}{\textbf{Mean}}  & 6356.56            & 1589.17           \\
\multicolumn{2}{|c|}{\textbf{Median}} & 6363.00            & 1590.50           \\ 
\multicolumn{2}{|c|}{\textbf{Min}}   & 6019               & 1505              \\ 
\multicolumn{2}{|c|}{\textbf{Max}}   & 6677               & 1669              \\ 
\multicolumn{2}{|c|}{\textbf{Standard deviation}}   & 164.73             & 41.25             \\ \hline
\end{tabular}

\end{table}

The dataset includes both raw and pre-processed data. Raw sensor data is organized per participant, with each entry capturing subject ID, activity label, timestamp, and tri-axial sensor readings (x, y, z). These raw logs are segmented by device and sensor type and stored in a structured directory format.

For the purposes of this research, we exclusively utilized the raw accelerometer data from this dataset. Gyroscope readings, smartwatch data, and transformed features were excluded to maintain consistency with our sensor modality focus across datasets.

The 18 activities recorded in the dataset represent a broad spectrum of daily and recreational movements: walking, jogging, stair climbing, sitting, standing, typing, brushing teeth, eating soup, eating chips, eating pasta, drinking from a cup, eating a sandwich, kicking a soccer ball, playing catch with a tennis ball, dribbling a basketball, writing, clapping, and folding clothes.

\subsubsection{Evaluation Metrics}
\label{experiment:evaluation_metrics}

Model performance is evaluated using a set of core metrics that reflect both overall effectiveness and class-specific reliability.

\begin{itemize}
    \item \textbf{Overall Accuracy} and \textbf{F1-Score} provide a general view of the model's performance. While accuracy shows the proportion of correct predictions across all classes, F1-Score balances precision and recall, making it more reliable in cases of class imbalance.
    \item \textbf{Per-class metrics} help assess how well the model handles each label, these collectively reveal the model's confidence and error tendencies for each class:
    \begin{itemize}
        \item \textbf{Sensitivity} indicates the ability to correctly identify true positives.
        \item \textbf{PPV (Precision)} shows how many predicted positives are actually correct.
        \item \textbf{NPV} reflects how many predicted negatives are truly negative.
    \end{itemize}
    
\end{itemize}

This combination of metrics offers both a broad and detailed understanding of model behavior, especially in imbalanced classification tasks.

\subsection{Experimental Design}
\label{experiment_design}

\subsubsection{Data Preprocessing}
The three-dimensional acceleration data (x, y, z axes) from all files were concatenated and labeled according to the corresponding posture (or activity in the MHealth dataset) and sensor placement location. The data points were subsequently transformed into sequences using a sliding window approach with a sequence length of 100 and a step size of 60. The most frequently occurring label within each sequence was assigned as the label for the entire sequence.

The data were then reshaped to the format (no\_samples, 1, 3, sequence\_length) to ensure compatibility with the CNN model architecture employed in this study. The processed dataset was partitioned into training and testing sets in an 80:20 ratio. The training data were further divided into five folds to facilitate cross-validation during model training.

\subsubsection{Method Comparison}

\textbf{Single-Task Learning (Baseline)}: This foundational approach trains two separate MTL-net models independently, with each model dedicated to a single task using standard Cross-Entropy loss. One model focuses exclusively on activity recognition while the other predicts device location, serving as the primary baseline for comparison.

\textbf{Multi-Task Learning (Baseline)}: A unified MTL-net model is trained simultaneously on both tasks using a composite loss function that represents the weighted average of two Cross-Entropy losses, controlled by parameter $\alpha$ as defined in Equation \ref{eq:multitask_loss}.

\textbf{Self-Distillation via Dropout (SD-Dropout)}: Following the methodology proposed by Lee et al. \cite{lee2022selfknowledgedistillationdropout}, this approach applies dropout with probability 0.5 to create diverse feature views from identical input representations. Knowledge distillation is performed between these views using a total loss formulation that applies distillation separately to each task, following the same principle as equation \ref{eq:total_loss} with balanced weighting parameter $\alpha$.

\textbf{Conventional Self-Distillation (Born-Again SD \cite{born_again_SD})}: This traditional two-stage knowledge distillation method serves as a performance benchmark, with detailed procedures outlined in section \ref{sec:conventional_born_again_sd}. The loss function has been adapted for multi-task learning using the proposed loss function in Equation \ref{eq:total_loss}, similar to the proposed method. The approach utilizes temperature parameter $\tau = 3.0$ and distillation weight $\lambda = 0.5$.

\textbf{Smooth-Distill (Proposed)}: The proposed online knowledge distillation framework employs a dynamically smoothed teacher model. The complete methodology is detailed in section \ref{sec:method}, with optimized parameters $\lambda = 0.5$, $\beta = 0.999$, and distillation temperature 3.0.

\subsubsection{Architectural, Data, and Hyperparameter Variants for Ablation Study}
\label{ex:architecture_data_hyperparameter}

\paragraph{Model Architecture and Training Ratio}
To comprehensively evaluate the effectiveness of our proposed approach, we conducted ablation studies examining both model architecture variations and training data sensitivity. To analyze the impact of model architecture, we performed comparative analysis across four different neural network architectures including LSTM, Bidirectional LSTM, GRU, and AnpoNet, alongside our previously proposed MTL-Net architecture. Each architecture was evaluated under identical experimental conditions to isolate the influence of architectural design choices on multi-task learning performance.

\begin{itemize}
    \item \textbf{LSTM Model:} A 2-layer recurrent neural network that processes sequential data through LSTM cells with gating mechanisms to capture long-term dependencies. The architecture includes dropout regularization and shared feature extraction layers before task-specific classification heads.
    \item \textbf{Bidirectional LSTM Model:} A 2-layer bidirectional recurrent architecture that processes sequences in both forward and backward directions to capture temporal dependencies from past and future contexts. Hidden states from both directions are concatenated and processed through shared layers for enhanced sequence modeling.
    \item \textbf{GRU Model:} A simplified 2-layer recurrent architecture using Gated Recurrent Units with fewer parameters than LSTM. The model employs gating mechanisms without separate cell states, providing computational efficiency while capturing temporal dependencies through shared feature extraction layers.
    \item \textbf{AnpoNet:} A hybrid architecture proposed by Vu et al. \cite{VuHoangDieu2025} for time series data, combining convolutional layers with batch normalization for initial feature extraction and LSTM layers for temporal modeling. AnpoNet integrates the spatial learning capabilities of CNNs and the temporal dynamics of RNNs through a unified feature extraction framework.
\end{itemize}

To analyze sensitivity to data partitioning, we conducted experiments with varying training ratios to assess model robustness across different data availability scenarios. The training process excluded cross-validation methodology, instead utilizing a fixed train/validation/test split with the training portion varied systematically from 10\% to 90\% of the available training data. Following training and validation phases, the model weights achieving the highest average validation accuracy across both tasks were selected for final evaluation on the test set. All model architectures were trained using the proposed Smooth-Distill methodology to ensure consistent experimental conditions.

\paragraph{Hyperparameter Sensitivity}
To assess the sensitivity of our proposed Smooth-Distill method to the knowledge distillation weight $\lambda$, we conducted a systematic analysis across a range of values including 0.001, 0.1, 0.5, and 1.0. The weight $\lambda$ serves as a coefficient in the total loss function, balancing the cross-entropy loss and the knowledge distillation loss, as detailed in Equation \ref{eq:total_loss}. This experiment was performed using the MTL-Net architecture with the Smooth-Distill training methodology to evaluate the impact of distillation weight on overall system performance and training stability.

\subsubsection{Implementation Details}

All experimental implementations utilized PyTorch framework with Adam optimizer configured at learning rate 0.001 and batch size 64. Training proceeded for 300 epochs on NVIDIA A100 GPUs. The experimental protocol maintained $\alpha = 0.5$ across all multi-task configurations to ensure balanced task weighting.

Each method underwent evaluation using five independent random seeds with results averaged across all runs. Model selection was performed based on average validation accuracy across both tasks, with optimal weights subsequently evaluated on the reserved test set. Five-fold cross-validation with identical fold divisions ensured fair comparison across all methodologies, providing robust performance assessment while minimizing initialization and partitioning effects.

\section{Results}
\label{sec:results}

\begin{table*}[t]
\centering 
\caption{Comparison of cross-validation and test accuracies for five training methods: Singletask, Multitask, Multitask with Self-Distillation via Dropout \cite{lee2022selfknowledgedistillationdropout} (SD-Dropout), Multitask with Born-Again Self-Distillation \cite{born_again_SD} (Born-Again SD), and our proposed Multitask with Smooth-Distillation (Smooth-Distill). The models are evaluated on the multitasks of posture/activity recognition (Task 1) and device location identification (Task 2) across three distinct accelerometer datasets.}
\label{tab:accuracy_comparison}
\setlength{\tabcolsep}{3pt} 
\resizebox{1.0\textwidth}{!}{%
\begin{tabular}{@{}lc 
                      cc 
                      cc 
                      cc@{}} 
\toprule
& \multirow{2}{*}{Methods} 
& \multicolumn{2}{c}{Sleep}
& \multicolumn{2}{c}{MHealth}
& \multicolumn{2}{c}{WISDM} \\
\cmidrule(lr){3-4} \cmidrule(lr){5-6} \cmidrule(lr){7-8}
& 
& Accuracy & F1-Score
& Accuracy & F1-Score
& Accuracy & F1-Score \\
\midrule
\multirow{5}{*}{Task1 val} & Singletask & $89.51 \pm 0.33$ & $89.48 \pm 0.34$ & $74.63 \pm 0.51$ & $51.03 \pm 0.65$ & $63.84 \pm 0.89$ & $63.79 \pm 0.86$ \\
& Multitask & $89.53 \pm 1.02$ & $89.50 \pm 1.03$ & $74.71 \pm 0.53$ & $50.20 \pm 1.00$ & $64.26 \pm 0.84$ & $64.17 \pm 0.79$ \\
& SD-Dropout & $87.39 \pm 1.39$ & $87.36 \pm 1.42$ & $75.56 \pm 0.49$ & $48.60 \pm 1.95$ & $61.14 \pm 2.02$ & $61.15 \pm 1.92$ \\
& Born-Again SD & $92.00 \pm 0.22$ & $91.99 \pm 0.21$ & $77.68 \pm 0.15$ & $50.68 \pm 2.04$ & $66.84 \pm 0.08$ & $66.84 \pm 0.13$ \\
& Smooth-Distill & $\mathbf{92.07 \pm 0.58}$ & $\mathbf{92.06 \pm 0.59}$ & $\mathbf{78.34 \pm 0.10}$ & $\mathbf{55.22 \pm 1.00}$ & $\mathbf{68.45 \pm 0.38}$ & $\mathbf{68.41 \pm 0.35}$ \\
\midrule
\multirow{5}{*}{Task1 test} & Singletask & $89.74 \pm 0.38$ & $89.70 \pm 0.39$ & $74.64 \pm 0.52$ & $49.89 \pm 1.19$ & $64.05 \pm 0.71$ & $64.02 \pm 0.71$ \\
& Multitask & $90.09 \pm 1.14$ & $90.05 \pm 1.15$ & $74.79 \pm 0.41$ & $49.28 \pm 1.02$ & $64.46 \pm 0.69$ & $64.40 \pm 0.67$ \\
& SD-Dropout & $87.66 \pm 1.31$ & $87.61 \pm 1.31$ & $75.72 \pm 0.51$ & $47.68 \pm 1.56$ & $61.36 \pm 2.08$ & $61.38 \pm 2.01$ \\
& Born-Again SD & $91.96 \pm 0.11$ & $91.93 \pm 0.12$ & $77.62 \pm 0.25$ & $48.87 \pm 1.60$ & $66.80 \pm 0.25$ & $66.83 \pm 0.28$ \\
& Smooth-Distill & $\mathbf{92.22 \pm 0.53}$ & $\mathbf{92.19 \pm 0.53}$ & $\mathbf{78.35 \pm 0.14}$ & $\mathbf{53.97 \pm 0.96}$ & $\mathbf{68.59 \pm 0.33}$ & $\mathbf{68.58 \pm 0.32}$ \\
\midrule
\multirow{5}{*}{Task2 val} & Singletask & $92.08 \pm 1.18$ & $91.80 \pm 1.25$ & $95.89 \pm 0.22$ & $95.87 \pm 0.23$ & $88.42 \pm 0.41$ & $88.24 \pm 0.45$ \\
& Multitask & $91.50 \pm 1.05$ & $91.20 \pm 1.09$ & $95.58 \pm 0.21$ & $95.57 \pm 0.21$ & $88.50 \pm 0.52$ & $88.33 \pm 0.51$ \\
& SD-Dropout & $88.72 \pm 1.02$ & $88.28 \pm 1.08$ & $95.11 \pm 0.30$ & $95.09 \pm 0.30$ & $86.78 \pm 0.68$ & $86.58 \pm 0.70$ \\
& Born-Again SD & $93.23 \pm 0.55$ & $93.02 \pm 0.58$ & $95.98 \pm 0.25$ & $95.96 \pm 0.26$ & $88.35 \pm 0.21$ & $88.14 \pm 0.21$ \\
& Smooth-Distill & $\mathbf{93.38 \pm 0.74}$ & $\mathbf{93.15 \pm 0.76}$ & $\mathbf{96.41 \pm 0.22}$ & $\mathbf{96.39 \pm 0.23}$ & $\mathbf{89.24 \pm 0.31}$ & $\mathbf{89.06 \pm 0.32}$ \\
\midrule
\multirow{5}{*}{Task2 test} & Singletask & $91.78 \pm 1.46$ & $91.50 \pm 1.52$ & $95.79 \pm 0.21$ & $95.79 \pm 0.21$ & $88.53 \pm 0.31$ & $88.32 \pm 0.34$ \\
& Multitask & $91.57 \pm 1.08$ & $91.29 \pm 1.13$ & $95.37 \pm 0.19$ & $95.37 \pm 0.18$ & $88.51 \pm 0.47$ & $88.32 \pm 0.46$ \\
& SD-Dropout & $88.85 \pm 0.90$ & $88.42 \pm 0.98$ & $95.13 \pm 0.44$ & $95.12 \pm 0.44$ & $86.94 \pm 0.79$ & $86.71 \pm 0.83$ \\
& Born-Again SD & $93.23 \pm 0.41$ & $93.00 \pm 0.44$ & $95.86 \pm 0.09$ & $95.85 \pm 0.09$ & $88.37 \pm 0.17$ & $88.14 \pm 0.19$ \\
& Smooth-Distill & $\mathbf{93.40 \pm 0.58}$ & $\mathbf{93.18 \pm 0.63}$ & $\mathbf{96.27 \pm 0.19}$ & $\mathbf{96.26 \pm 0.19}$ & $\mathbf{89.05 \pm 0.41}$ & $\mathbf{88.85 \pm 0.44}$ \\
\bottomrule
\end{tabular}
}
\end{table*}

\subsection{Overall Performance Comparison}

The final results of our study are presented in Table \ref{tab:accuracy_comparison}, which compares the accuracy and F1-scores on validation and test sets for two classification tasks: activity recognition and device placement position. We evaluated five methods: Singletask training, Multitask training, Self-Distillation via Dropout (SD-Dropout), Born-Again Self-Distillation (Born-Again SD), and our proposed Smooth-Distill method. The evaluation was conducted using accelerometer data from three datasets: one Sleep dataset collected by our team and two publicly available HAR datasets previously described in Section \ref{experiment:dataset}. Our proposed Smooth-Distill method consistently outperformed other approaches across nearly all tasks and datasets.

The results demonstrate the effectiveness and stability of our proposed method compared to baseline approaches, particularly evident in the WISDM dataset, where we achieved 68.59\% test accuracy for Task 1 and 89.05\% for Task 2. These results represent improvements of approximately 4\% for Task 1 and 0.5\% for Task 2 compared to baseline methods, outperforming all other approaches in both activity recognition and device placement position identification.

For datasets with significant label imbalance, such as task 1 of the MHealth dataset, the disparity between F1-score and accuracy metrics was substantially larger compared to other datasets. Our proposed method maintained its effectiveness under these challenging conditions, achieving the highest performance among all evaluated approaches. The performance gap between the proposed method and competing approaches was even more pronounced when evaluated using F1-scores, highlighting the robustness of our approach to class imbalance.

In cases involving more easily classifiable data, such as sleep postures in the Sleep dataset or device placement position classification, the improvements were more modest and could be considered equivalent in some instances. However, when considering the overall performance across both tasks simultaneously, the cumulative improvement represents a substantial enhancement.

The conventional self-distillation approach (Born-Again SD) also achieved strong performance, outperforming most other methods and demonstrating performance comparable to our proposed method. However, there still remains a large F1-score gap in HAR classification on the MHealth dataset compared to the proposed method.

The SD-Dropout method showed underperformance, with results that were predominantly lower than or equivalent to Singletask and Multitask approaches across most tasks and datasets. This finding suggests that the dropout-based distillation strategy may not be optimal for the accelerometer-based classification tasks examined in this study.

Singletask training demonstrated underperformance across all datasets. This approach can also be considered more resource-intensive, requiring separate training of two distinct models for each task, yet the results failed to meet expectations, performing only equivalently to or worse than basic multitask training approaches.

\subsection{Detailed Performance Metrics}

\begin{table*}[t]
\centering
\caption{Comparison of performance metrics for five methods: Singletask, Multitask, Self-Distillation via Dropout \cite{lee2022selfknowledgedistillationdropout} (SD-Dropout), Born-Again Self-Distillation \cite{born_again_SD} (Born-Again SD), and Smooth-Distill.}
\label{tab:details_performance_metrics}
\setlength{\tabcolsep}{3pt}
\resizebox{\textwidth}{!}{%
\begin{tabular}{@{}llcccccccccccc@{}}
\toprule
\multicolumn{2}{c}{Label} & 0 & 1 & 2 & 3 & 4 & 5 & 6 & 7 & 8 & 9 & 10 & 11 \\
\midrule
\multirow{5}{*}{Acc} & Singletask & $98.67$ & $98.28$ & $98.66$ & $98.82$ & $\mathbf{98.64}$ & $98.57$ & $98.41$ & $97.63$ & $98.24$ & $98.33$ & $98.32$ & $98.01$ \\
& Multitask & $98.37$ & $97.86$ & $98.13$ & $98.68$ & $97.95$ & $98.10$ & $98.24$ & $97.26$ & $98.28$ & $98.06$ & $97.86$ & $97.54$ \\
& SD-Dropout & $98.90$ & $98.40$ & $98.49$ & $98.94$ & $98.49$ & $98.54$ & $98.80$ & $98.18$ & $98.39$ & $98.19$ & $97.93$ & $97.97$ \\
& Born-Again SD & $98.92$ & $\mathbf{98.78}$ & $\mathbf{98.93}$ & $\mathbf{99.12}$ & $98.56$ & $98.64$ & $98.68$ & $98.15$ & $98.78$ & $98.63$ & $98.69$ & $\mathbf{98.52}$ \\
& Smooth-Distill & $\mathbf{98.98}$ & $98.62$ & $98.80$ & $\mathbf{99.12}$ & $98.63$ & $\mathbf{98.78}$ & $\mathbf{98.82}$ & $\mathbf{98.31}$ & $\mathbf{98.84}$ & $\mathbf{98.79}$ & $\mathbf{98.77}$ & $98.35$ \\
\midrule
\multirow{5}{*}{Sen} & Singletask & $94.48$ & $91.55$ & $91.10$ & $92.73$ & $93.26$ & $90.50$ & $92.55$ & $84.62$ & $89.63$ & $88.97$ & $89.38$ & $84.32$ \\
& Multitask & $87.50$ & $88.65$ & $89.46$ & $90.45$ & $89.66$ & $90.95$ & $87.43$ & $82.05$ & $89.11$ & $90.39$ & $82.30$ & $\mathbf{89.91}$ \\
& SD-Dropout & $93.69$ & $90.77$ & $87.24$ & $93.70$ & $93.03$ & $92.08$ & $\mathbf{92.88}$ & $87.06$ & $88.90$ & $91.27$ & $87.94$ & $88.36$ \\
& Born-Again SD & $\mathbf{95.27}$ & $\mathbf{92.88}$ & $91.80$ & $\mathbf{94.57}$ & $91.69$ & $\mathbf{93.21}$ & $92.32$ & $86.36$ & $92.46$ & $92.58$ & $\mathbf{94.25}$ & $88.36$ \\
& Smooth-Distill & $95.05$ & $91.43$ & $\mathbf{93.44}$ & $94.25$ & $\mathbf{93.82}$ & $92.65$ & $92.77$ & $\mathbf{88.69}$ & $\mathbf{92.77}$ & $\mathbf{92.79}$ & $93.03$ & $87.65$ \\
\midrule
\multirow{5}{*}{PPV} & Singletask & $90.02$ & $88.40$ & $92.07$ & $93.54$ & $90.61$ & $92.06$ & $88.98$ & $85.61$ & $90.58$ & $91.27$ & $90.58$ & $89.76$ \\
& Multitask & $92.39$ & $86.26$ & $87.41$ & $94.02$ & $86.18$ & $86.73$ & $91.18$ & $83.51$ & $91.41$ & $87.34$ & $91.51$ & $80.96$ \\
& SD-Dropout & $\mathbf{93.06}$ & $90.27$ & $93.36$ & $93.91$ & $89.22$ & $90.44$ & $92.78$ & $89.89$ & $92.79$ & $88.00$ & $87.56$ & $86.21$ \\
& Born-Again SD & $91.96$ & $\mathbf{92.57}$ & $\mathbf{94.57}$ & $95.19$ & $\mathbf{91.07}$ & $90.55$ & $92.02$ & $\mathbf{90.15}$ & $93.74$ & $91.48$ & $90.64$ & $\mathbf{92.42}$ \\
& Smooth-Distill & $92.85$ & $92.05$ & $91.62$ & $\mathbf{95.49}$ & $90.08$ & $\mathbf{92.54}$ & $\mathbf{93.18}$ & $90.06$ & $\mathbf{94.16}$ & $\mathbf{93.00}$ & $\mathbf{92.42}$ & $91.00$ \\
\midrule
\multirow{5}{*}{NPV} & Singletask & $99.50$ & $\mathbf{99.22}$ & $99.23$ & $99.32$ & $99.39$ & $99.15$ & $99.31$ & $98.66$ & $98.99$ & $98.97$ & $99.02$ & $98.67$ \\
& Multitask & $98.88$ & $98.96$ & $99.09$ & $99.10$ & $99.06$ & $99.18$ & $98.85$ & $98.44$ & $98.94$ & $99.10$ & $98.38$ & $\mathbf{99.13}$ \\
& SD-Dropout & $99.43$ & $99.15$ & $98.90$ & $99.41$ & $99.37$ & $99.29$ & $\mathbf{99.35}$ & $98.88$ & $98.92$ & $99.18$ & $98.89$ & $99.00$ \\
& Born-Again SD & $\mathbf{99.57}$ & $\mathbf{99.35}$ & $99.29$ & $\mathbf{99.49}$ & $99.25$ & $\mathbf{99.39}$ & $99.30$ & $98.82$ & $99.26$ & $99.30$ & $\mathbf{99.47}$ & $99.01$ \\
& Smooth-Distill & $99.55$ & $\mathbf{99.22}$ & $\mathbf{99.43}$ & $99.46$ & $\mathbf{99.44}$ & $99.34$ & $99.34$ & $\mathbf{99.02}$ & $\mathbf{99.29}$ & $\mathbf{99.33}$ & $99.36$ & $98.95$ \\
\bottomrule
\end{tabular}%
} 
\end{table*}

Additional performance metrics for the four methods applied to Task 1 of the Sleep dataset are presented in Table \ref{tab:details_performance_metrics}, showing accuracy, sensitivity, positive predictive value (PPV), and negative predictive value (NPV) for each label. This table reveals more distinct differences between methods, particularly in sensitivity and PPV metrics when comparing multitask and multitask smooth distillation approaches. Sensitivity measures the model's ability to correctly identify positive cases, while PPV indicates the proportion of positive predictions that are actually correct, both crucial for reliable classification performance.

Both Smooth-Distill and Born-Again SD demonstrated superior and stable performance compared to other methods across all metrics and most labels, with Smooth-Distill showing slight advantages while Born-Again SD remained competitive with comparable results. SD-Dropout showed improvements in several metrics compared to baseline methods.

Although the overall accuracy results provided in Table \ref{tab:accuracy_comparison} indicated that the multitask method did not lag significantly behind our proposed method, the detailed metrics reveal substantially lower performance for multitask training, particularly in sensitivity and PPV across multiple labels. This demonstrates the clear improvement achieved by our proposed method over standard multitask approaches. The enhanced performance at the class level indicates better discrimination capability and more reliable predictions across different sleep postures.

\begin{figure*}[t]%
\centering
\includegraphics[width=1.0\textwidth]{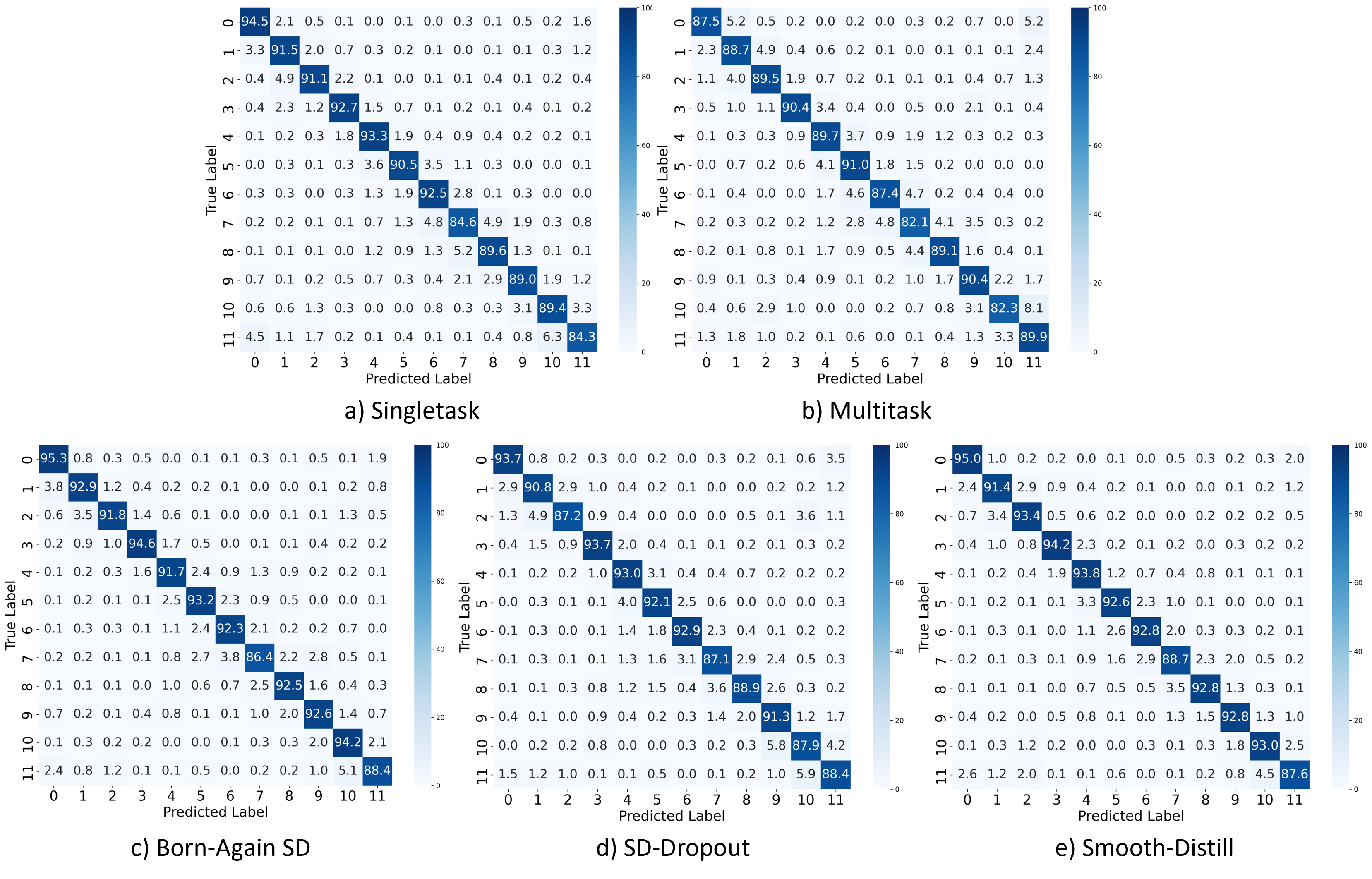}
\caption{Confusion Matrices for Task 1 (Posture/Activity Recognition) of Five Methods on the Sleep Dataset.}
\label{fig:confusion_matrices_figure}
\end{figure*}

The confusion matrices for these five methods on the Sleep dataset are illustrated in Figure \ref{fig:confusion_matrices_figure}. The results show that most misclassifications occur between adjacent labels, which in the Sleep dataset correspond to sleep postures differing by only 30 degrees of rotation. This represents a common challenge with this dataset, which includes numerous sleep postures that create difficulties not only during model prediction but also during data collection. Nevertheless, these results demonstrate that all models can handle this dataset effectively, with the Smooth-Distill method showing modest advantages over other approaches.

Despite these inherent dataset challenges, the results demonstrate that all models possess adequate capability to handle this complex classification task, with our Smooth-Distill method showing a slight but consistent advantage. The confusion matrix patterns confirm that the model errors are logical and primarily occur between semantically similar classes, suggesting that the learned representations capture meaningful relationships between different sleep postures while achieving improved discrimination through our proposed distillation approach.

\subsection{Ablation Studies and Sensitivity Analysis}
\subsubsection{Impact of Knowledge Distillation Weight ($\lambda$)}

\begin{figure*}[t]%
\centering
\includegraphics[width=0.9\textwidth]{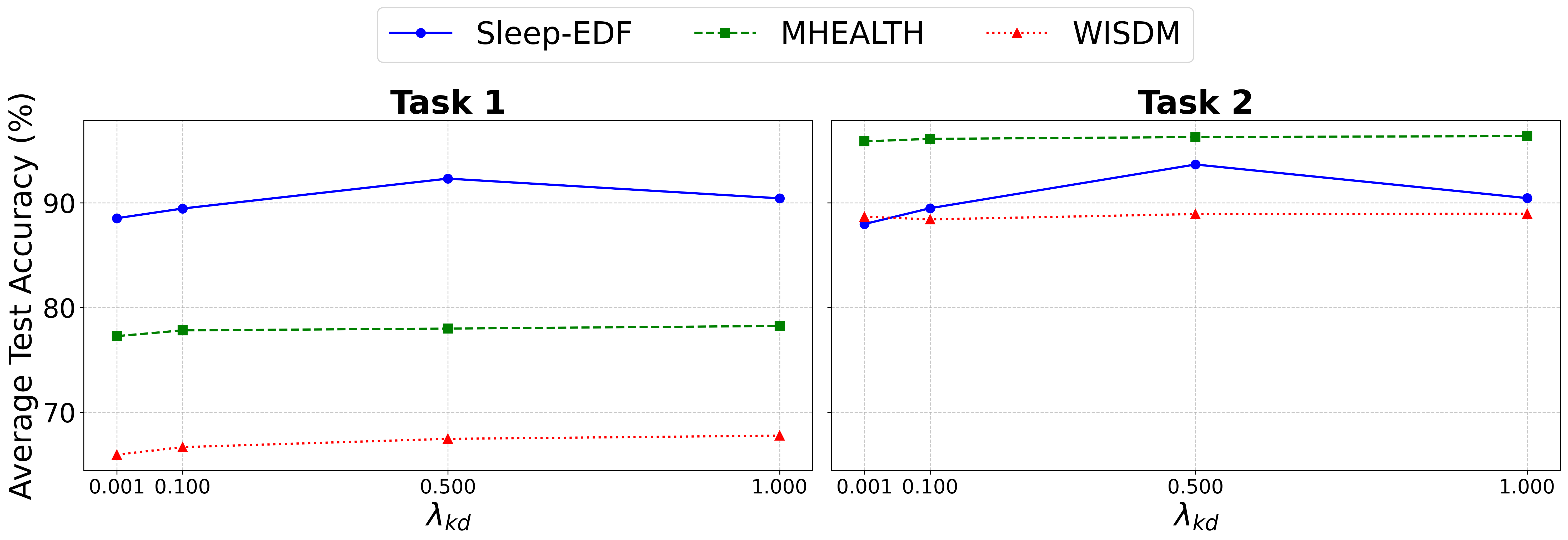}
\caption{Test Accuracy of Multitask Smooth-Distill for Task 1 (Posture/Activity Recognition) and Task 2 (Device Location Identification) with Varying Lambda Values Across Three Datasets (Sleep, MHealth, WISDM).}
\label{fig:lambda_vs_acc}
\end{figure*}

As outlined in Section \ref{ex:architecture_data_hyperparameter}, we investigated the influence of the knowledge distillation weight $\lambda$ on the performance of Smooth-Distill. Figure \ref{fig:lambda_vs_acc} presents the results of this analysis, showing the test accuracy on all three datasets as $\lambda$ varies.

The experimental findings indicate that $\lambda$ values of 0.5 and 1.0 provide superior optimization, with $\lambda = 0.5$ demonstrating particularly exceptional performance enhancement for the Sleep dataset, substantially outperforming other parameter settings. While $\lambda = 1.0$ shows marginal improvements for the MHealth and WISDM datasets, the performance differences remain modest. Given that $\lambda = 0.5$ produces remarkable improvements for the Sleep dataset while maintaining competitive performance across other datasets, this value represents the optimal configuration for the evaluated data domains.

\subsubsection{Analysis of Model Architectures and Training Ratios}

\begin{figure*}[t]%
\centering
\includegraphics[width=0.9\textwidth]{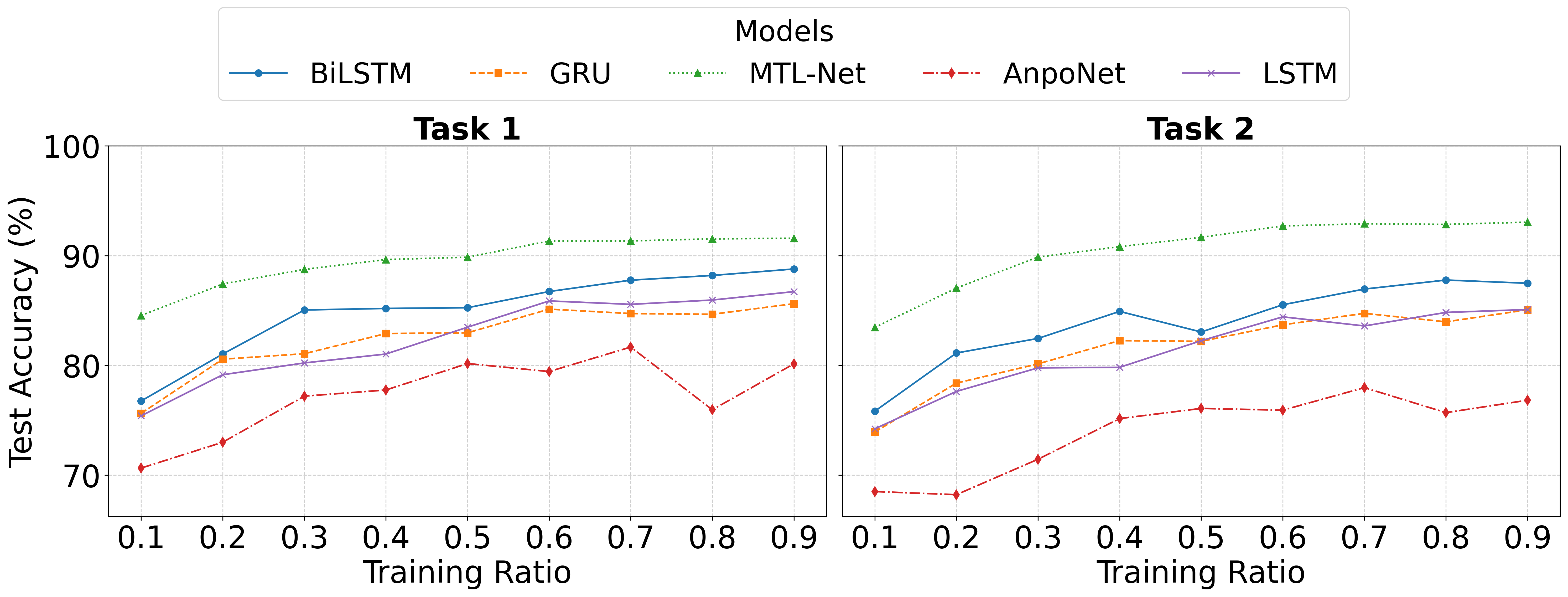}
\caption{Comparative Test Accuracy of Model Architectures on the Sleep Dataset Across Varying Training Ratios.}
\label{fig:trainratio_sleep}
\end{figure*}

\begin{figure*}[t]%
\centering
\includegraphics[width=0.9\textwidth]{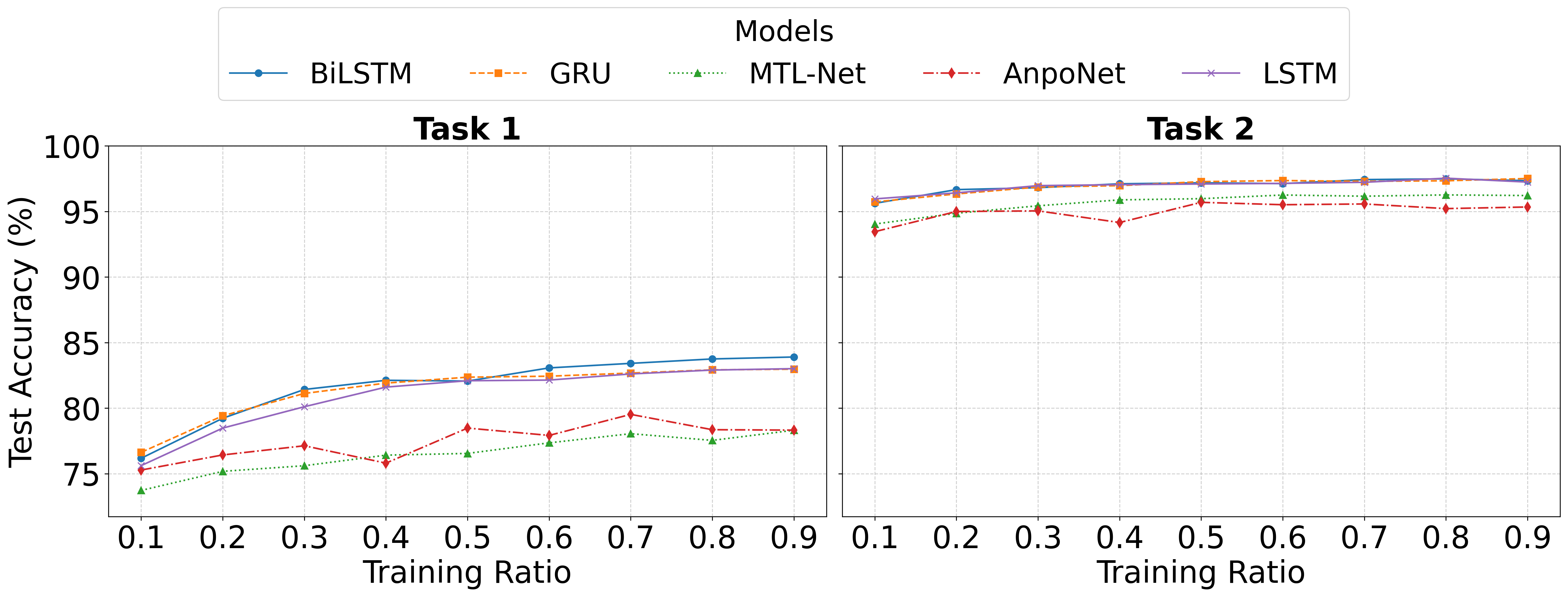}
\caption{Comparative Test Accuracy of Model Architectures on the MHealth Dataset Across Varying Training Ratios.}
\label{fig:trainratio_mhealth}
\end{figure*}

\begin{figure*}[t]%
\centering
\includegraphics[width=0.9\textwidth]{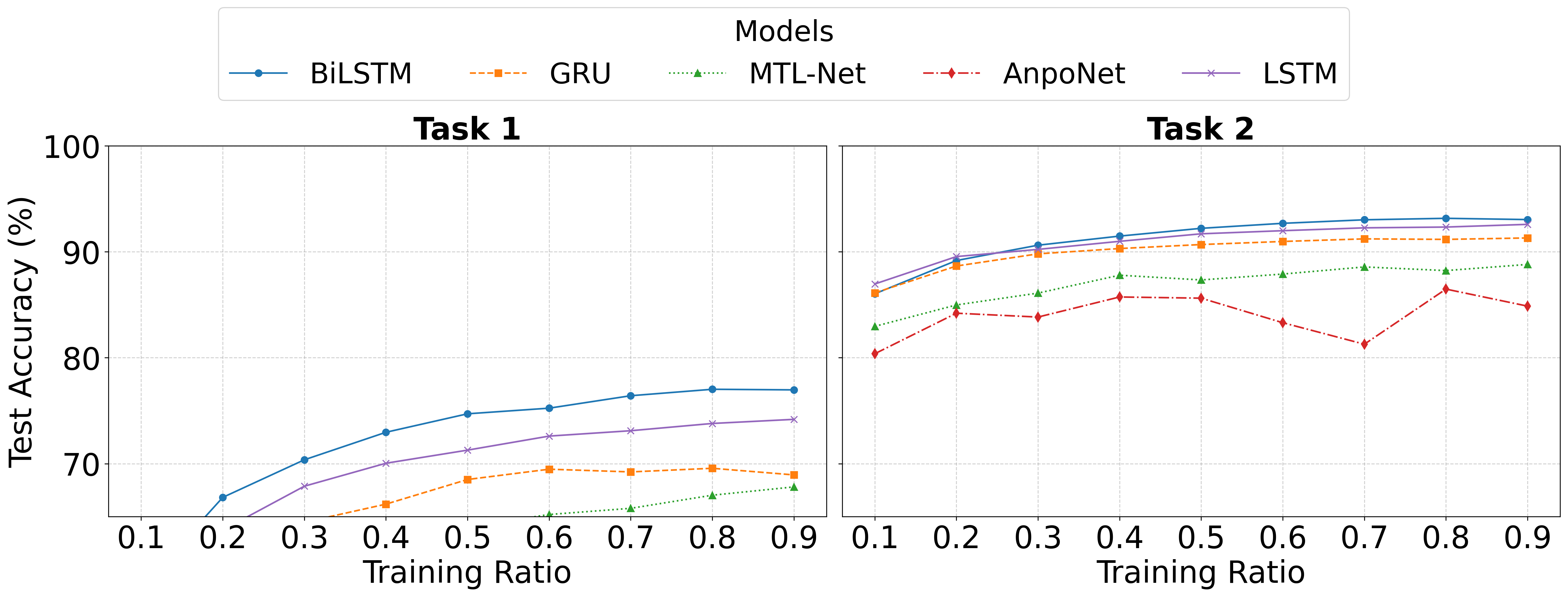}
\caption{Comparative Test Accuracy of Model Architectures on the WISDM Dataset Across Varying Training Ratios.}
\label{fig:trainratio_wisdm}
\end{figure*}

We further analyze how different model architectures and training data ratios influence Smooth-Distill's effectiveness across datasets. The comparative results of five models—LSTM, BiLSTM, GRU, MTL-Net, and AnpoNet—are presented across Figures \ref{fig:trainratio_sleep}--\ref{fig:trainratio_wisdm}, corresponding to the Sleep, MHealth, and WISDM datasets respectively.

The MTL-Net model achieved the highest accuracy on the Sleep dataset (Figure \ref{fig:trainratio_sleep}), which represents a relatively simple dataset where acceleration data between classes exhibits clear separation without requiring consideration of acceleration fluctuations for differentiation. This characteristic prevents RNN structures from fully leveraging their capabilities, thereby enabling CNN-based architectures such as MTL-Net to maximize their effectiveness and to achieve optimal performance on this dataset.

RNN-based models, including LSTM, BiLSTM, GRU, and AnpoNet architectures, typically demonstrate considerably higher computational complexity compared to CNN-based models. However, these models excel when strong temporal dependencies exist between data points within a time series window. This advantage is particularly evident in human activity recognition datasets such as MHealth (Figure \ref{fig:trainratio_mhealth}) and WISDM (Figure \ref{fig:trainratio_wisdm}), where activity classification requires careful analysis of both the intensity and frequency variations in acceleration signals. On these datasets, RNN models consistently outperform the MTL-Net architecture, with BiLSTM specifically demonstrating superior performance compared to other RNN structures across most datasets and training ratios.

The AnpoNet, which combines CNN and LSTM components, failed to demonstrate satisfactory results across all three datasets (Figures \ref{fig:trainratio_sleep}--\ref{fig:trainratio_wisdm}), frequently achieving the lowest performance while simultaneously exhibiting high sensitivity to training ratio variations. This suboptimal performance could be attributed to the architectural complexity that may lead to optimization difficulties and the potential mismatch between the CNN feature extraction phase and the subsequent LSTM temporal modeling, resulting in inefficient information flow and degraded learning capacity.

Regarding training ratio analysis, models maintained effective classification performance for both human activity recognition and sensor placement tasks using only 10\% of the training data, demonstrating the robustness of the proposed approach under limited data conditions. The results reveal a consistent trend where test accuracy gradually increases with higher training ratios, with most models reaching an elbow point at approximately 70\% training ratio. This pattern suggests that while additional training data beyond this threshold provides marginal improvements, the 70\% ratio represents an optimal balance between data utilization efficiency and model performance for the evaluated tasks.

\section{Discussion}
\label{sec:discussion}

\subsection{Analysis of Computational Trade-offs and Practicality}

\begin{table}[t]
\centering
\caption{Comparison of Model Characteristics, Training Efficiency, and Energy Cost for All Methods on the Sleep Dataset.}
\label{tab:model_characteristics}
\sisetup{group-digits=true, group-separator={,}}

\begin{tabular}{@{}l S[table-format=5.3] S[table-format=1.3] S[table-format=8.0] S[table-format=5.0] S[table-format=2.3]@{}}
\toprule
Method          & {Size (MB)} & {Parameters (M)} & {FLOPS}     & {Training Time (s)} & {Energy Cost (kWh)} \\
\midrule
Singletask      & \num{13.710} & \num{3.592}      & \num{10010112} & \num{5012}            & \num{0.557} \\
Multitask       & \num{6.862}  & \num{1.798}      & \num{5008128}  & \num{3878}            & \num{0.431} \\
SD-Dropout \cite{lee2022selfknowledgedistillationdropout} & \num{6.862}  & \num{1.798}      & \num{5008128}  & \num{13001}           & \num{1.444} \\
Born-Again SD \cite{born_again_SD} & \num{6.862}  & \num{1.798}      & \num{5008128}  & \num{8280}            & \num{0.920} \\
Smooth-Distill  & \num{6.862}  & \num{1.798}      & \num{5008128}  & \num{4368}            & \num{0.485} \\
\bottomrule
\end{tabular}
\end{table}

Beyond classification accuracy, the practical deployment of machine learning models requires a careful analysis of their computational trade-offs. In this section, we discuss the computational characteristics of our proposed Smooth-Distill method in comparison to other approaches. While Smooth-Distill introduces a moderate computational overhead for its self-distillation mechanism, it establishes a favorable balance between the investment in training resources and the significant gains in performance, making it a viable solution for sensor-based applications.

These efficiency advantages are demonstrated comprehensively in Table~\ref{tab:model_characteristics}, which presents model size, parameter count, FLOPS, training time, and energy cost for each method evaluated in our research. The energy cost is calculated based on the power consumption of the NVIDIA A100 GPU and the training time required for each model.

Singletask training (for one task) is used as the baseline in our comparisons. This choice provides several advantages for comparative analysis. Singletask training represents the most conventional and widely adopted training paradigm in machine learning applications, making it an appropriate reference point for evaluation. This baseline selection enables more intuitive performance comparisons since features traverse the same network layers in both singletask and multitask architectures, with the primary distinction occurring at the output branching stage. This structural similarity facilitates direct assessment of the computational and performance impact when transitioning from singletask to multitask learning frameworks.

Given that this research addresses multitask learning scenarios requiring classification across two distinct tasks, the conventional singletask approach necessitates training two separate models, each dedicated to one specific classification objective. This methodology results in substantially increased computational costs, effectively doubling all resource requirements including model size, parameters, FLOPS, and training time compared to our established baseline. However, despite this significant computational investment, the performance outcomes for the dual classification tasks examined in this research demonstrate suboptimal results relative to the resource expenditure.

Multitask training introduces additional computational overhead that, while relatively modest in absolute terms, results in approximately 1.5 times longer training duration compared to the baseline. However, this approach remains significantly more efficient than the conventional alternative while delivering superior performance across both classification tasks.

Our proposed method increases training time to approximately 1.72 times the baseline, representing a reasonable computational investment that delivers markedly superior performance improvements. This balance between computational efficiency and performance enhancement positions Smooth-Distill as a practical solution for real-world applications.

The SD-Dropout method, despite offering storage advantages, demonstrates significant limitations in training efficiency. The approach requires dropout vector features to provide multiple model views for mutual distillation, resulting in training times approximately 5.18 times longer than the baseline, making it the most computationally expensive method evaluated. Despite its promising theoretical foundation, the substantial time investment yields disappointing results, suggesting that this dropout-based distillation strategy may be better suited for image or video data as demonstrated in the original research \cite{lee2022selfknowledgedistillationdropout}, rather than time series or accelerometer data applications.

Born-Again SD delivers relatively strong performance comparable to our proposed method, yet requires pre-training a separate teacher model that increases total training time to approximately 3.3 times the baseline. This computational expense makes it a relatively costly training approach that achieves equivalent or slightly inferior performance compared to our proposed method, highlighting the efficiency advantages of Smooth-Distill.

Energy consumption analysis reveals significant operational implications across the evaluated methods. The multitask approach demonstrates the highest energy efficiency at 0.431 kWh, achieving a 23\% reduction compared to the singletask baseline of 0.557 kWh, while SD-Dropout exhibits the poorest efficiency at 1.444 kWh, consuming 2.6 times more energy than the baseline. Our proposed Smooth-Distill method achieves 0.485 kWh energy consumption, representing a 13\% improvement over the baseline and positioning it as the second most efficient approach while delivering superior performance compared to multitask training. This energy efficiency advantage translates to meaningful operational cost savings and reduced environmental impact in large-scale deployments where models undergo frequent retraining cycles.

\subsection{Statistical Significance Analysis}

\begin{table*}[t]
\caption{Statistical Significance Tests (p-values) for pairwise comparison of five methods on sleep posture recognition task using Sleep dataset.}
\centering

\begin{subtable}{\textwidth}
\centering
\setlength{\tabcolsep}{4pt}
\resizebox{0.75\textwidth}{!}{%
\begin{tabular}{|l|c|c|c|c|c|}
\hline
 & \textbf{Singletask} & \textbf{Multitask} & \textbf{SD-Dropout} & \textbf{Born-Again SD} & \textbf{Smooth-Distill} \\
\hline
\textbf{Singletask} & 1 & 0.0421 & 0.4631 & 0.0003 & 0.0001 \\
\hline
\textbf{Multitask} & 0.0421 & 1 & 0.0099 & 0.0011 & 0.0003 \\
\hline
\textbf{SD-Dropout} & 0.4631 & 0.0099 & 1 & 0.0005 & 0.0006 \\
\hline
\textbf{Born-Again SD} & 0.0003 & 0.0011 & 0.0005 & 1 & 0.7173 \\
\hline
\textbf{Smooth-Distill} & 0.0001 & 0.0003 & 0.0006 & 0.7173 & 1 \\
\hline
\end{tabular}
}
\subcaption{Paired t-tests Comparing Mean Fold Accuracy}
\label{tab:accuracy_pvalue}
\end{subtable}

\vspace{1em} 

\begin{subtable}{\textwidth}
\centering
\resizebox{0.75\textwidth}{!}{%
\begin{tabular}{|l|c|c|c|c|c|}
\hline
 & \textbf{Singletask} & \textbf{Multitask} & \textbf{SD-Dropout} & \textbf{Born-Again SD} & \textbf{Smooth-Distill} \\
\hline
\textbf{Singletask} & 1 & 0.0418 & 0.4420 & 0.0003 & 0.0001 \\
\hline
\textbf{Multitask} & 0.0418 & 1 & 0.0091 & 0.0011 & 0.0004 \\
\hline
\textbf{SD-Dropout} & 0.4420 & 0.0091 & 1 & 0.0005 & 0.0006 \\
\hline
\textbf{Born-Again SD} & 0.0003 & 0.0011 & 0.0005 & 1 & 0.7431 \\
\hline
\textbf{Smooth-Distill} & 0.0001 & 0.0004 & 0.0006 & 0.7431 & 1 \\
\hline
\end{tabular}
}
\subcaption{Paired t-tests Comparing Mean Fold F1 Score}
\label{tab:f1_pvalue}
\end{subtable}

\label{tab:p_values_comparison}
\end{table*}

The comparative analysis of p-values presented in Table \ref{tab:p_values_comparison} reveals a clear stratification of performance among the evaluated methods. The results demonstrate two distinct performance tiers, with Born-Again SD and Smooth-Distill forming the top tier and exhibiting statistically indistinguishable performance characteristics ($p > 0.7$). This finding suggests that both methods achieve comparable levels of effectiveness, establishing them as the most robust approaches within the evaluated framework.

The performance hierarchy becomes particularly evident when examining cross-tier comparisons. The top-tier methods demonstrate statistically significant superiority over all bottom-tier approaches, including Singletask, Multitask, and SD-Dropout methods ($p < 0.001$ across all comparisons). This substantial performance gap underscores the effectiveness of the advanced distillation techniques employed by Born-Again SD and Smooth-Distill. Within the bottom tier, the results reveal limited effectiveness of certain approaches, with SD-Dropout failing to provide statistically significant improvement over the Singletask baseline ($p > 0.44$), indicating that dropout-based regularization offers minimal benefit in this context. Meanwhile, standard Multitask learning demonstrates only marginal gains over the Singletask approach ($p \approx 0.04$), suggesting that while some improvement is achieved, the enhancement remains relatively modest compared to the substantial advances offered by the top-tier methods.

\subsection{Task Complexity Considerations and Multitask Learning Benefits}

The relative difficulty difference between Task 1 and Task 2 presents important implications for method comparison. Task 2 demonstrates considerably easier classification characteristics compared to Task 1, which could substantially alter results if both tasks exhibited similar difficulty levels, potentially favoring single-task approaches when both tasks are equally challenging. Nevertheless, the obtained results remain valuable since simultaneous prediction of both posture or activity and device placement position addresses practical requirements, particularly relevant for future work involving optimization based on device positioning.

The easier nature of Task 2 relative to Task 1 creates favorable conditions for multitask learning approaches over single-task alternatives. Multitask methods generally achieve superior performance while significantly reducing computational costs and training time compared to training two separate single-task models. This efficiency advantage explains why the majority of methods developed in this study employ multitask architectures rather than single-task approaches.

\subsection{Training Convergence Analysis}

\begin{figure*}[t]%
\centering
\includegraphics[width=1.0\textwidth]{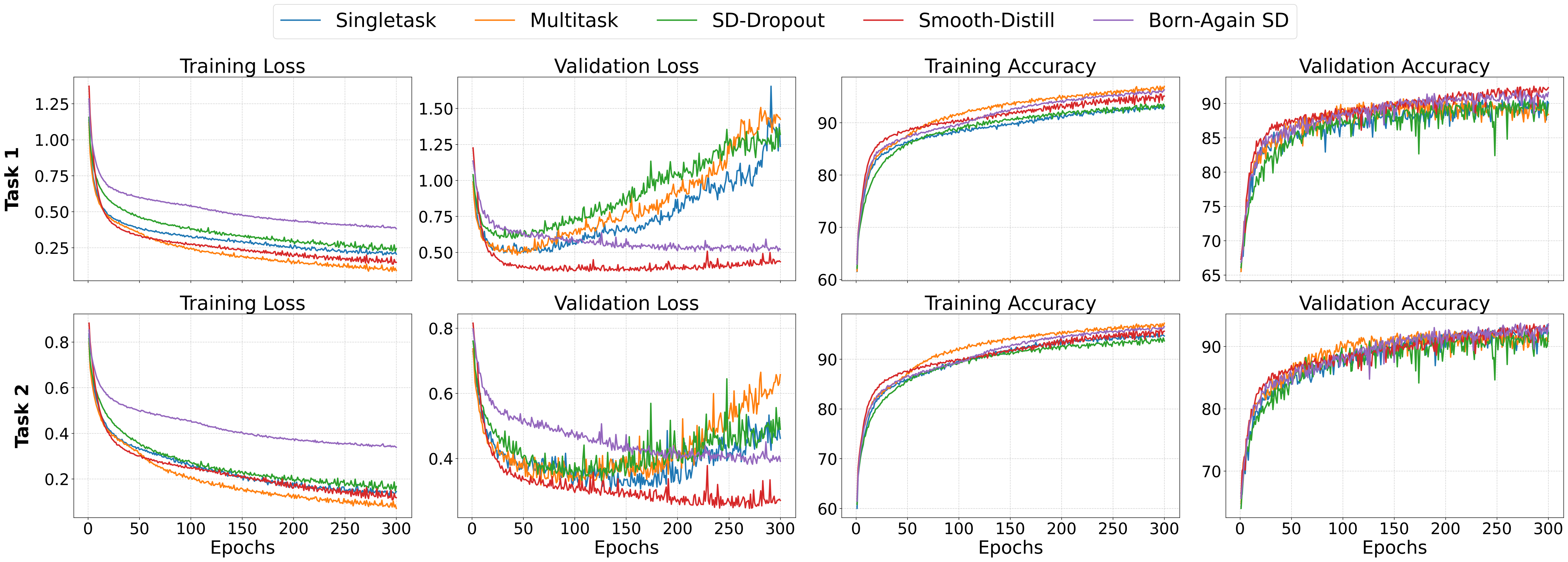}
\caption{Training Curves of Four Methods for Posture/Activity Recognition (Task 1) and Device Location Identification (Task 2) on the Sleep Dataset.}
\label{fig:training_curves}
\end{figure*}

The training curves for all five methods are illustrated in Figure \ref{fig:training_curves}, displaying training loss, training accuracy, validation loss, and validation accuracy across both tasks of the Sleep dataset. These curves reveal the substantial benefits of self-distillation approaches, particularly our proposed Smooth-Distill method, in facilitating more stable and reliable model convergence. During initial epochs, all methods demonstrate similar convergence rates, establishing comparable learning trajectories across different approaches.

However, as training progresses to later epochs, significant differences emerge in model behavior. Baseline methods increasingly exhibit overfitting characteristics, with training loss continuing to decrease sharply while validation loss begins to increase, indicating poor generalization to unseen data. In contrast, both Born-Again SD and Smooth-Distill maintain more balanced learning dynamics, with training loss decreasing while validation loss remains stable or continues to improve, and validation accuracy continuing to increase across both classification tasks.

The training curves also demonstrate the superiority of Smooth-Distill over Born-Again SD through distinct convergence patterns. Born-Again SD exhibits relatively slow training loss reduction compared to other methods, which contributes to maintaining generalization capability and allows validation loss to continue decreasing while baseline methods begin to overfit. Our proposed Smooth-Distill method shows only slightly slower training loss reduction during initial epochs but subsequently achieves rapid convergence similar to baseline methods while maintaining consistent validation loss improvement. Notably, Smooth-Distill achieves the fastest and lowest validation loss reduction among all evaluated methods.

These convergence characteristics demonstrate the scalability potential of our approach, enabling faster and more reliable model convergence that establishes a foundation for handling larger datasets and more complex model architectures. The method's ability to achieve superior performance improvements over traditional approaches without excessive computational overhead makes it a practical solution for accelerometer-based classification tasks. This convergence analysis supports the conclusion that Smooth-Distill represents a reasonable and effective approach for scenarios requiring enhanced model performance while maintaining computational efficiency.

\subsection{Implications for Time Series Classification}

The comprehensive evaluation demonstrates that our Smooth-Distill approach addresses fundamental challenges in multitask time series classification, particularly for accelerometer-based applications. The method successfully balances computational efficiency with performance enhancement, providing a practical solution for scenarios requiring simultaneous classification of multiple related tasks. The stable convergence characteristics and consistent performance improvements across diverse datasets suggest broad applicability for similar time series classification problems.

While our method shows consistent improvements, the magnitude of enhancement varies depending on dataset characteristics and task complexity. For datasets with easily separable classes---such as the Sleep dataset or device placement detection tasks---performance gains may be more modest, though the cumulative effect across multiple tasks remains substantial. 

Future research directions should encompass several key areas. Foremost, examining the framework's adaptability to sophisticated neural architectures and extensive datasets would establish its scalability limits. Furthermore, applying the methodology to alternative time series domains beyond accelerometry—including gyroscopic, magnetometric, or physiological signals—would demonstrate cross-modal effectiveness. Critically, constructing a more robust and methodologically rigorous original dataset represents a priority for future work. Such a resource, incorporating stricter validation protocols and expanded coverage of operational scenarios, would facilitate comprehensive evaluation of both algorithmic performance and computational resource demands. This enhanced benchmark would enable more precise characterization of the method's efficacy under varying real-world constraints and usage patterns.

\section{Conclusion}
\label{sec:conclusion}

This research presents Smooth-Distill, a novel self-distillation approach for accelerometer-based classification tasks that addresses the need for efficient multitask learning in wearable sensor applications. Through comprehensive experimentation across three datasets, our method demonstrates consistent performance improvements while maintaining computational efficiency.

The experimental results show clear performance advantages for Smooth-Distill across multiple evaluation metrics. The method achieved notable improvements of approximately 4\% in activity recognition and 0.5\% in device placement classification on the WISDM dataset, with consistent superior performance across all evaluated datasets. These improvements are particularly significant for challenging scenarios involving imbalanced datasets, where traditional approaches often struggle to maintain reliable classification performance.

The computational efficiency analysis demonstrates that Smooth-Distill achieves optimal balance between performance enhancement and resource utilization. While the method requires approximately 72\% more training time than individual singletask models due to distillation loss calculation and dynamic teacher model updates, this represents a reasonable computational investment that remains substantially more efficient than training separate models or employing more complex distillation strategies like Born-Again SD or SD-Dropout. The hyperparameter analysis establishes $\lambda = 0.5$ as the optimal configuration for balancing cross-entropy and distillation loss components.

The training convergence analysis provides compelling evidence of the method's stability and scalability potential. Smooth-Distill demonstrates superior convergence characteristics, maintaining consistent validation performance while achieving faster convergence compared to baseline approaches. This behavior indicates better generalization capability and suggests strong potential for application to larger datasets and more complex model architectures.

Future research should investigate the scalability of Smooth-Distill with larger datasets, more complex model architectures, and diverse time series modalities, including gyroscopic and physiological signals. Evaluating its performance under different sampling settings and with enriched datasets—-featuring stricter validation and broader operational coverage-—will offer deeper insight into its generalizability, resource efficiency, and real-world applicability.

In conclusion, Smooth-Distill represents a significant advancement in multitask learning for accelerometer-based classification tasks, offering a practical solution that consistently delivers superior performance while maintaining computational feasibility. The method's reported effectiveness across diverse scenarios, along with its training-time efficiency, highlights its potential for widespread adoption in wearable sensor applications where model development costs are a concern.

\section*{Acknowledgement}
This work was supported by the Vietnam National Foundation for Science and Technology Development (NAFOSTED) under Grant No. NCUD.02-2024.11, and by the French National Research Agency (ANR) in the framework of the Investissements d’avenir program (ANR-10- AIRT-05 and ANR-15-IDEX-02), and the MIAI Cluster (ANR-23-IACL-0006). This work also forms part of a broader translational and interdisciplinary GaitAlps research program (NV).

\printbibliography

\end{document}